\documentclass{article}

\usepackage{arxiv}

\usepackage[utf8]{inputenc} 
\usepackage[T1]{fontenc}    
\usepackage{hyperref}       
\usepackage{url}            
\usepackage{booktabs}       
\usepackage{amsfonts}       
\usepackage{nicefrac}       
\usepackage{microtype}      
\usepackage{lipsum}
\usepackage[export]{adjustbox}

\usepackage[section]{placeins}
\usepackage[shortlabels]{enumitem}
\usepackage{subcaption}
\usepackage[numbers]{natbib}
\usepackage{amsmath,amssymb,amsfonts}
\usepackage{algorithmic}
\usepackage{graphicx}
\usepackage{textcomp}
\usepackage{xcolor}
\def\BibTeX{{\rm B\kern-.05em{\sc i\kern-.025em b}\kern-.08em
    T\kern-.1667em\lower.7ex\hbox{E}\kern-.125emX}}
    
\usepackage{verbatim}  

\usepackage{hyperref}
\usepackage{csquotes}
\usepackage{todonotes}
\usepackage{float}

\usepackage{ tipa }

\usepackage[utf8]{inputenc}

\title{A Unified Substrate for Body-Brain Co-evolution}

\author{Sidney Pontes-Filho$^{1,2,*}$, Kathryn Walker$^{3}$, Elias Najarro$^{3}$, Stefano Nichele$^{1,4,5}$ and Sebastian Risi$^{3}$
\bigskip \\ 
$^1$Department of Computer Science, Oslo Metropolitan University, Oslo (Norway)\\
$^2$Department of Computer Science, Norwegian University of Science and Technology, Trondheim (Norway)\\
$^3$Digital Design Department, IT University of Copenhagen, Copenhagen (Denmark)\\
$^4$Department of Holistic Systems, Simula Metropolitan Centre for Digital Engineering, Oslo (Norway)\\
$^5$Department of Computer Science and Communication, Østfold University College, Halden (Norway)\\
$^*$Corresponding author: sidneyp@oslomet.no}

\date{\small}

\begin{document}
\maketitle
\begin{abstract}
The discovery of complex multicellular organism development took millions of years of evolution. The genome of such a multicellular organism guides the development of its body from a single cell, including its control system. Our goal is to imitate this natural process using a single neural cellular automaton (NCA) as a genome for modular robotic agents. In the introduced approach, called \emph{Neural Cellular Robot Substrate} (NCRS), a single NCA guides the growth of a robot and the cellular activity which controls the robot during deployment. We also introduce three benchmark environments, which test the ability of the approach to grow different robot morphologies. In this paper, NCRSs are trained with covariance matrix adaptation evolution strategy (CMA-ES), and covariance matrix adaptation MAP-Elites (CMA-ME) for quality diversity, which we show leads to more diverse robot morphologies with higher fitness scores. While the NCRS can solve the easier tasks from our benchmark environments, the success rate reduces when the difficulty of the task increases. We discuss directions for future work that may facilitate the use of the NCRS approach for more complex domains.
\end{abstract}

\section{Introduction}

Multicellular organisms are made of cells that can divide into many, which specialize in controlling and maintaining the body, sensing the environment, or protecting from external threats. Such features were acquired by evolution from the first living cell. After millions of years, colonies of unicellular organisms appeared and were essential to the development of multicellular organisms with cellular differentiation \citep{niklas2013origins}. Developmental biologists study that the growth and specialization of an organism are coordinated by its genetic code \citep{slack2021essential}.

The field of artificial life tries to create life-like computational models taking ideas from biological life, such as decentralized and local control \citep{langton2019artificial}. One of the sub-fields of artificial life, artificial development \citep{harding2009artificial,doursat2013morpho}, focuses on modeling or simulating cell division and differentiation. The techniques applied in artificial development are often based on the indirect encoding of developmental rules (i.e.\ analogous to the genome of a biological organism describing its phenotype). This type of encoding facilitates the scaling of an organism because the information in the genome is much smaller than in the resulting phenotype. This property is referred to as genomic bottleneck \citep{zador2019critique,variengien}, and it implies that the genetic code of an organism compresses the information to grow and maintain its body, and in some species even complex brains.

One of the simplest computational models of artificial life or dynamical systems is a cellular automaton (CA) \citep{wolfram2002new}. A CA can be described as a universe with discrete space and time, which is governed by local rules without any central control. Such a discrete space is divided into a regular grid of cells and can possess any number of dimensions. The most commonly studied CAs have one or two dimensions and their most well-known versions are, respectively, elementary CA \citep{wolfram2002new} and Conway’s Game of Life \citep{conway1970game}. Both have cells with binary states, but other CA can have many discrete states or continuous ones. In the 1940s, the first CA was introduced by Ulam and von Neumann \citep{topa2011network}. Von Neumann aimed to produce self-replicating machines, and Ulam worked on crystal growth. In 2002, a CA with rules defined by an artificial neural network was described \citep{li2002neural}. Nowadays, this type of approach is called neural cellular automaton (NCA). In 2017, \citet{nichele2017neat} presented an NCA that has developmental features that were learned through neuroevolution using a method called compositional pattern-producing network \citep{stanley2007compositional}. Recently, \citet{mordvintsev2020growing} introduced a differentiable NCA, which possesses growth and regeneration properties. In their work, an NCA is trained through gradient descent to grow a colored image from one active "seed" cell.

\begin{figure}[ht]
  \centering
  \includegraphics[width=0.9\textwidth]{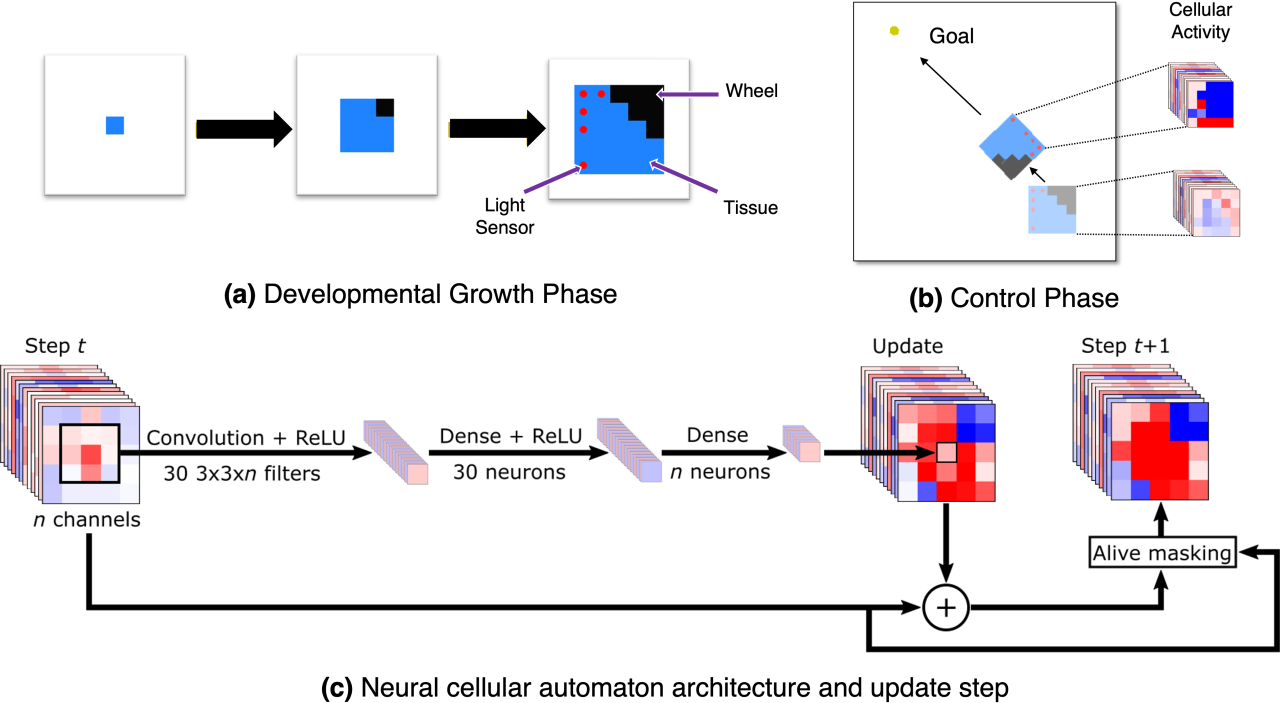}
  \caption{Neural Cellular Robot Substrate (NCRS) \normalfont In the developmental phase (a), the robot is grown from an initial starting seed, guided by a neural cellular automaton (c). Once grown, the same neural cellular automaton determines how the signals propagate through the robot's morphology during the control phase (b).}
  \label{fig:teaser}
\end{figure}

In evolutionary robotics, co-evolution of morphology and control has the inherent challenge of optimizing two different features in parallel \citep{bhatia2021evolution}. It also presents scalability issues when it deals with modular robots \citep{yim2007modular}. Our goal is to implement an approach where the optimization happens in just one dynamical substrate with local interactions. Here we introduce such a system, a \emph{Neural Cellular Robot Substrate} (NCRS), in which a single NCA grows the morphology of an agent’s body and also controls how that agent interacts with its environment. The NCA has two phases (Fig.~\ref{fig:teaser}). First is the developmental phase, in which the robot's body is grown, including where to place its sensors and actuators. In the following control phase, copies of the same NCA are running in each cell of the agent, taking into account only local information from neighboring cells to determine their next state. The optimization task thus entails figuring out how to transmit information from the robot's sensors to its actuators to perform the task at hand.

\begin{figure}[ht]
  \centering
  \subcaptionbox{Light chasing with obstacle task\label{fig:env2}}{%
    \includegraphics[width=0.5\linewidth,frame]{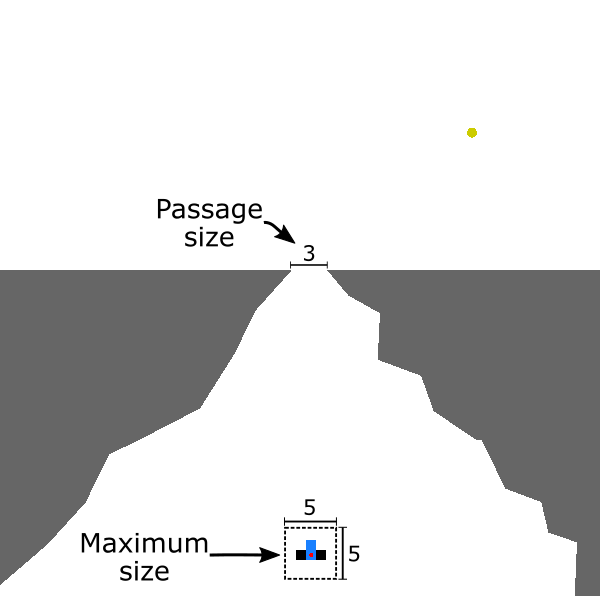}
  }\hspace{1.5cm}
  \subcaptionbox{Carrying ball to target task\label{fig:env3}}{%
    \includegraphics[width=0.35\linewidth,frame]{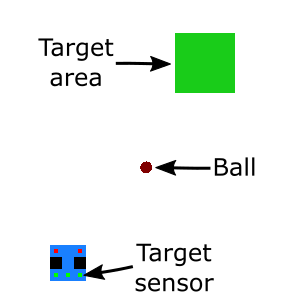}
  }
  \caption{Extensions from the light chasing task. (\subref{fig:env2}) It depicts the original size of the playing field, which is 60.}
  \label{fig:env23}
\end{figure}

We also introduce a virtual environment with three benchmark tasks for evaluating the NCRS' capacity of designing a robot and then controlling it. Two benchmarks consist in growing and controlling a robot to approach a light source (Fig.~\ref{fig:teaser}b and Fig.~\ref{fig:env2}). 
The third task challenges the robot to carry a ball to a target area. In this benchmark, a second type of rudimentary eye is added, so the robot can differentiate the ball and the target area (Fig.~\ref{fig:env3}). 

The main contribution of this work is the introduction of a single neural cellular automaton that first grows an agent's body and then controls it during deployment. While the solved benchmark domains are relatively simple, the unified substrate for both body and brain opens up interesting future research directions, such as opportunities for open-ended evolution \citep{stanley2019open}. The source code of this project and videos of the results are available at \url{https://github.com/sidneyp/neural-cellular-robot-substrate}.

\section{Related work}

The co-design of robot bodies and brains has been an active area of research for decades \citep{medvet2021biodiversity,sims1994evolving,komosinski1999framsticks,veenstra2020different,gupta2021embodied}. Brain and body co-design stands for producing a control policy and a morphology for a robotic system. 
For example, in the work of \citet{lipson2000automatic}  the same genome directly encodes the robot's body and the artificial neural network for control. A method that uses genetic regulatory networks to define separately a body and an artificial neural network was introduced by \citet{bongard2003evolving} and named artificial ontogeny. The evolved robots are able to locomote and push blocks in noisy environments. More recent work by \citet{bhatia2021evolution} presents several virtual environments and also an algorithm for brain and body co-design with separated description methods for the morphology and control. In comparison with NCRS, our co-design algorithm consists of only one neural cellular automaton. 

The work on NCAs by \citet{mordvintsev2020growing} is one of the first examples of self-organizing and self-repair systems that use differentiable models as rules for cellular automata. Before that, NCA models were typically optimized with genetic algorithms \citep{nichele2017neat}. After the work on growing NCA, other neural CAs were introduced, including methods optimized without differentiable programming. There exist other generative methods for growing 3D artifacts and functional machines \citep{sudhakaran2021growing}, for developing soft robots \citep{horibe2021regenerating}. Moreover, an NCA was used as a decentralized classifier of handwritten digits \citep{randazzo2020mnist}. 

The developmental phase of our approach is similar to the generative method with NCA for level design trained with CMA-ME in the work of \citet{earle2021illuminating}. Morphology design is also present in other works \citep{hejna2021task,talamini2021criticality,kriegman2018morphological,brodbeck2015morphological}. The control phase is based on the NCA for controlling a cart-pole agent introduced by \citet{variengien}, but their NCA is trained using a reinforcement learning algorithm named deep-Q learning and the communication between NCA and environment happens in predefined cells. Our approach, NCRS, unifies these two methods by having two phases. The first phase is generative, and the second one is an agent's policy.

\section{Approach: A Unified Substrate}

The modular robots grown by the NCA consist of different types of cells such as sensors, actuators, and connecting tissue. After growth, the robot is deployed in its particular environment. Importantly, in our approach, the same NCA controls both the growth of the modular robot (Fig.~\ref{fig:teaser}a) and the robot itself (Fig.~\ref{fig:teaser}b). Therefore, it is a unified substrate for body-brain co-design and is called Neural Cellular Robot Substrate (NCRS). The architecture of NCRS is illustrated in Fig.~\ref{fig:teaser}c. When the growth process is finished, the channels responsible to define the body modules reflect the robot's morphology, then the NCA can observe and act in the environment using the cells assigned to the specific types of modules, which are sensors, wheels, and tissue.

The state of a cell is updated considering the eight surrounding neighbors and itself, then it forms a $3\times 3$ neighborhood. The values of the nine cells with all the $n$ channels are processed by a trainable convolutional layer with 30 filters of size $3\times 3\times n$. Followed by a dense layer of 30 neurons and another one with $n$ neurons for the $n$ channels of the neural CA. After all cells have been computed, the result of this process is added to the previous state of the neural CA, and then it is clipped to the range of $[-5,5]$. This update is only valid for the cells that are considered "alive", which are the ones that have their value in the body channel greater than $0.1$ and their neighbors. This architecture is very similar to the ones in self-classifying MNIST \citep{randazzo2020mnist} and in self-organized control of a cart-pole agent \citep{variengien}.

\begin{figure}[tb]
\centering
  \subcaptionbox{Initial time-step in developmental phase\label{fig:channel_1}}{%
    \includegraphics[width=0.4\textwidth]{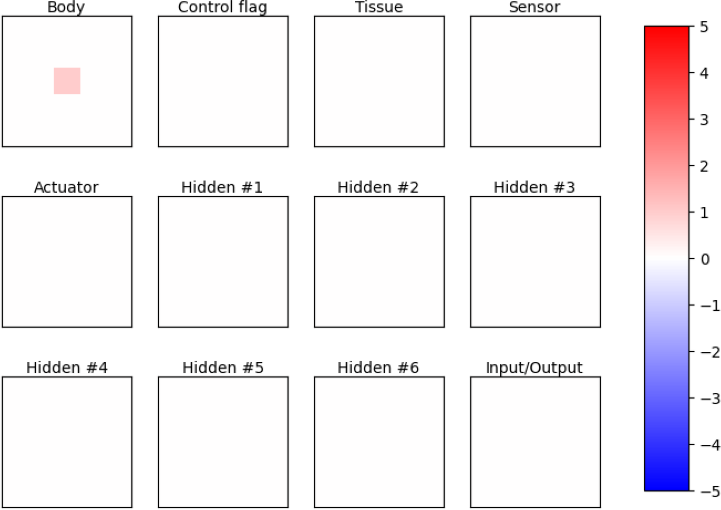}%
  }\hspace{1cm}
  \subcaptionbox{Final time-step of developmental phase\label{fig:channel_2}}{%
    \includegraphics[width=0.4\textwidth]{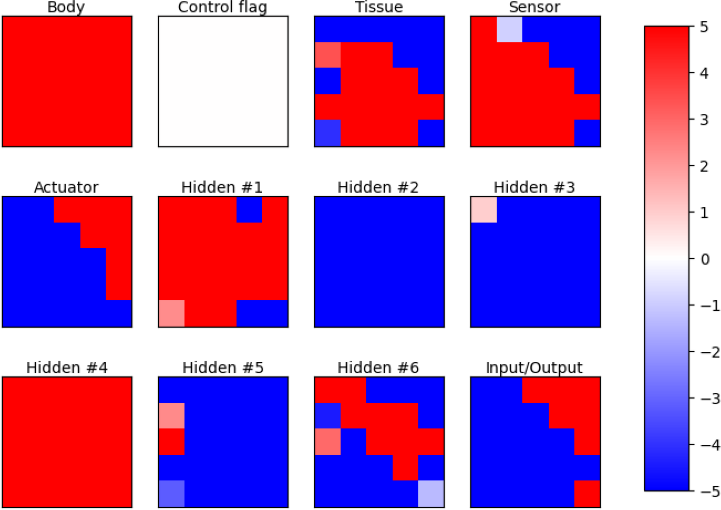}%
  }
  \subcaptionbox{Final time-step of control phase\label{fig:channel_3}}{%
    \includegraphics[width=0.4\textwidth]{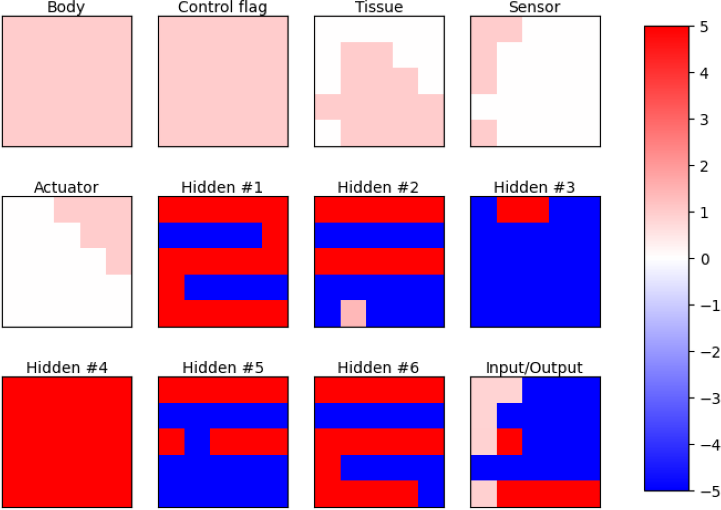}%
  }
  \caption{Channels of the neural cellular automaton in different stages.}
  \label{fig:channel}
\end{figure}

The channels have specific roles in the neural CA, as shown in Fig.~\ref{fig:channel}. The number of channels $n$ differs because of the different number of sensors in the types of benchmark tasks. The body channel is the one that indicates that there is a body part in that cell if its value is greater than $0.1$. The neighbors of a body part are allowed to update their states because they are considered "growing". The next channel has fixed values and works as a control flag. When the neural CA is in the developmental phase, all cells in this channel are set to zero. When it is in the control phase, they are set to one. The following channels are responsible to define the type of the body part. The channel with the highest value is the one that specifies the body part. In the case of a tie, the first channel is selected. The order of those channels is: body/brain tissue, light/ball sensor, target area sensor (if needed), and wheel. In this way, it can define a robot as depicted in Fig.~\ref{fig:teaser}a. Then, there are the hidden channels to support the computation in the neural CA. For all benchmark types, the neural CA contains six hidden channels. Finally, the input/output channel, which is the one that receives the values from the sensors and gives the values to the actuators (wheels).

The initial state of the neural CA is a "seed". The middle cell of the grid has the state set as one in the body channel, and the rest is zero. Fig.~\ref{fig:channel_1} illustrates this. After a few time-steps in the developmental phase, all channels are updated except the control flag channel. This phase lasts for ten time-steps. The end of the developmental phase is represented in Fig.~\ref{fig:channel_2}. After development, the control phase starts. In this phase, the benchmark environment initializes with the developed robot body. For advancing one time-step in the environment, the NCA takes two time-steps for defining an action after receiving observations from the sensors. The body and body parts channels become fixed and their values are defined by the robot body. This is used to support the neural CA by identifying the cells with body modules, such as tissues, sensors, and actuators. The cell is assigned the value one to the body channel if there is a body part and to the specific body parts channel. Fig.~\ref{fig:channel_3} shows this assignment for the identification of body parts during the control phase. The robot designed by this NCA is depicted in Fig.~\ref{fig:env1_es_body}. At the start of the control phase, the cellular activity of the hidden and input/output channels is set to zero. In the input/output channel, only the input cells are fixed and their values come from the sensors.

In our neural CA, there is no noise. Even all "alive" cells are updated every time-step. This is done because the stochastic update or any other type of noise would affect the development of the robot body. After the developmental phase, the same model could produce different types of robot body.

For our experiments, the neural cellular automaton has a grid of size $5\times 5$. Therefore, it generates a body for an agent with the same size. Since this is a neural CA, the grid size does not affect the number of trainable parameters. The light chasing and light chasing with obstacle environments require just the light sensor. Therefore, the robot can have tissue, light sensor and wheel. A wheel's orientation is always vertical during the initialization of the benchmark environment. The wheel rotates upwards and downwards relative to the initial angle of the robot. The maximum speeds for each of those directions are, respectively, +1 and -1. 
This takes three body part channels. With one body, one control flag, six hidden, and one input/output channels, the total number of channels is $12$. In this way, the number of trainable parameters is 4,572. In the carrying ball to target environment, the robot needs one additional sensor. Therefore, it adds one more channel. It results in a neural CA with 4,873 trainable parameters.

\section{Benchmark environments}

To test the capacity of controlling the developed robot, we implemented three benchmark environments, which are: light chasing (LC), light chasing with obstacle (LCO), and carrying ball to target (CBT). They are environments where a modular robot equipped with simple light sensors and wheels can be evaluated. In those environments, we decided that the size of the playing field and the distance between the objects are affected by the maximum size that the robot can have. Thus, the larger the robot can be, the bigger the playing field. In our experiments, we use a robot and a neural cellular automaton grid with size $5\times 5$. Because the possible maximum size of the robot is 5, we chose the size of the playing field to be 60.

The fitness score is calculated using the average score of 12 runs where the location of the agent, light, ball, and target can differ for each run. The light or ball has some predefined regions to be initially placed.

The benchmark environments are based on the implementation of the top-down racing environment in Open AI gym \citep{openaigym}. We use the pybox2d, which is a 2D physics library in Python. 

\subsection{Light chasing}

The light chasing (LC) environment is shown in Fig.~\ref{fig:teaser}b. The goal of the agent is to be closer to the light during the entire simulation. The agent starts in the middle of the playing field. One light is randomly placed around the region of one of the four corners of the playing field. The fitness score is calculated by the average distance between the center of the robot and the center of the light over all simulation time-steps, and a successful run means that this distance reached less than 10 times the module size. The activity $s$ of the agent's light sensors is calculated as: 
\begin{equation}
  s=e^{-distance/playfield}, 
  \label{eq:score}
\end{equation}
where the distance between the objects is normalized by the size of the playing field $playfield$, which is 60. 
The values of the sensor activity or fitness score are between 0 and 1, where 1 means no distance. The values exponentially decay to 0 with an increase in distance.

\subsection{Light chasing with obstacle}

The light chasing with obstacle (LCO) environment is a more difficult version of the light chasing one (Fig.~\ref{fig:env2}). 
The robot does not have sensors to detect the obstacle, thus its morphology plays a bigger role in this benchmark. The passage width is calculated by the possible maximum size of the robot. If the robot can have up to $5\times 5$ body parts, then the passage width would be the size of three body parts. 
The robot is randomly initialized at the bottom of the playing field. An obstacle is procedurally generated with a target passage width and wall roughness. The obstacle has the shape of a funnel because there are no sensors to it, then this helps the robot to reach the passage depending on its body. The passage is randomly located on the horizontal axis and fixed on the vertical axis. The light is at the top and after the obstacle. The initial light location has four predefined regions on the horizontal axis, which are left, center-left, center-right and right. The fitness and success definition are the same as the light chasing task.

\subsection{Carrying ball to target area}

Among the three benchmark environments, the task to carry a ball to a target area is the most difficult one (Fig.~\ref{fig:env3}).  
For the control phase, the robot needs to move towards the ball, and then move to the target area without losing the ball during the transport. For the developmental phase, the body of the robot needs to be adequate to push or kick the ball to the target area, and properly placing the sensors of each type, so it can successfully locate ball and target area. The agent is located at the bottom in a random horizontal location. The ball is located in the middle of the vertical axis of the playing field, but it has the same four predefined regions as the light chasing with obstacle environment. The target is located at the top and its location on the horizontal axis is randomly defined. Besides the sensor for the ball (or light for the other two environments), there is a new sensor type that calculates the distance to the center of the target area (following \eqref{eq:score}).

The fitness score of this environment is the average of the distance between robot and ball, and the distance between the ball and the center of the target area. Since they are distances used to calculate the fitness score, they are normalized using \eqref{eq:score}. The definition of success in this task means carrying the ball to the target, so it can have a distance less than ten times the module size of the robot.

\section{Training methods}

We have chosen to use some derivative-free optimization methods because NCRS needs some adjustments for using deep reinforcement learning because of the variable number of inputs and outputs \citep{variengien}. They are the covariance matrix adaptation evolution strategy (CMA-ES) \citep{hansen1996cmaes} and covariance matrix adaptation MAP-Elites (CMA-ME) \citep{fontaine2020cmame}. The latter is used to add quality diversity to the former, broadening the exploration of robot designs. For both training methods, we use the library CMA-ES/pycma \citep{hansen2019cma}. There are two training methods and three benchmark tasks. This gives a total of six different combinations. Because of the computational demands, each of these combinations was trained only once. 

The training process is performed entirely on a CPU. To speed up evaluation times, robots with a design that would not work properly in the environment are not simulated. For the two light chasing environments, robots must have at least one light sensor and two actuators. For the carrying ball to target, they must have one sensor of each type and two actuators. The fitness scores of the failed designs are calculated according to the number of correct parts they have. For each correct body part, the fitness score increases by $0.01$.

To compare the quality diversity of CMA-ES and CMA-ME, we use the percentage of cells or feature configurations filled, and QD-score. They measure quality and diversity of the elites \citep{pugh2016quality}. The QD-score is calculated by summing the fitness score of all elites and dividing it by the total number of possible feature configurations. Moreover, CMA-ES and CMA-ME have their elites stored, even though CMA-ES does not use elites during training.

\subsection{Covariance matrix adaptation evolution strategy}

CMA-ES is one of the most effective derivative-free numerical optimization methods for continuous domains \citep{fontaine2020cmame}. CMA-ES runs 20,000 generations for all environments. The initial mean is $0.0$ for all dimensions, and the initial coordinate-wise standard deviation (step size) is $0.01$. The population size or the number of solutions acquired to update the covariance matrix is 112. This number was selected by the number of available threads in the machine used to train, which contains 56 threads at 2.70GHz.

\subsection{Covariance matrix adaptation MAP-Elites}

CMA-ME is a variant of CMA-ES with the added benefit of quality diversity from MAP-Elites \citep{mouret2015mapelites}. The changes to CMA-ES are that there are emitters of CMA-ES being trained in a cycle. Additionally, a feature map stores one elite for each possible feature configuration. Because there are invalid body designs, they do not produce an elite. When there is a successful robot design, the number of sensors, actuators, and body parts are used as features. If there are no elites or the current solution is better than the actual elite stored in the feature map, then the current solution is assigned to its feature configuration.

We use a slightly modified version of the CMA-ME with improvement emitters \citep{fontaine2020cmame}. We only restart an emitter when the number of elites is greater than the number of emitters and it is stuck for more than 500 generations. Being stuck means that the emitter could not find a better elite or an elite could not be placed into an empty feature configuration in the map. When an emitter restarts, the mean used to initialize the CMA-ES is a random elite in the map.

CMA-ME is executed for 60,000 generations for all environments, except the light chasing environment with 67,446 generations because we forced it to stop a longer training and its best fitness score was already better than the one trained with CMA-ES. The initial mean and the initial coordinate-wise standard deviation are the same as CMA-ES for all emitters. The population size is 128 because the CMA-ME training was executed in a computer with 128 threads at 2.9GHz.

\section{Results}

The training process took around 2.5 days for optimizing the NCA with CMA-ES. The evolution with CMA-ME took around 5.5 days. It is important to note that they do not have the same machine configuration, population size, and maximum number of generations.

\begin{figure}[tpb]
\centering
  \subcaptionbox{CMA-ES - LC\label{fig:env1_es_body}}{%
    \includegraphics[width=0.14\textwidth]{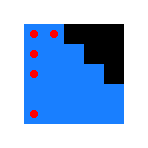}%
  }\hfill
  \subcaptionbox{CMA-ES - LCO\label{fig:env2_es_body}}{%
    \includegraphics[width=0.14\textwidth]{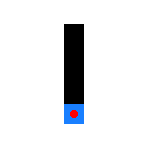}%
  }\hfill
  \subcaptionbox{CMA-ES - CBT\label{fig:env3_es_body}}{%
    \includegraphics[width=0.14\textwidth]{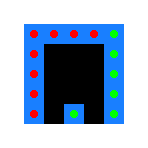}%
  }\hfill
  \subcaptionbox{CMA-ME - LC\label{fig:env1_cmame_body}}{%
    \includegraphics[width=0.14\textwidth]{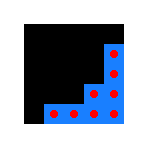}%
  }\hfill
  \subcaptionbox{CMA-ME - LCO\label{fig:env2_cmame_body}}{%
    \includegraphics[width=0.14\textwidth]{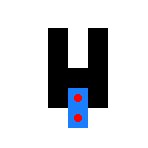}%
  }\hfill
  \subcaptionbox{CMA-ME - CBT\label{fig:env3_cmame_body}}{%
    \includegraphics[width=0.14\textwidth]{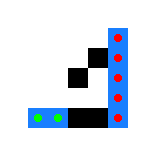}%
  }
  \caption{Robot designs with best fitness scores for the tasks of light chasing (LC), light chasing with obstacle (LCO) and carrying ball to target (CBT).}
  \label{fig:env_es_body}
\end{figure}

\begin{figure}[tpb]
\centering
  \subcaptionbox{Light chasing\label{fig:loss_env1}}{%
    \includegraphics[width=0.3\textwidth]{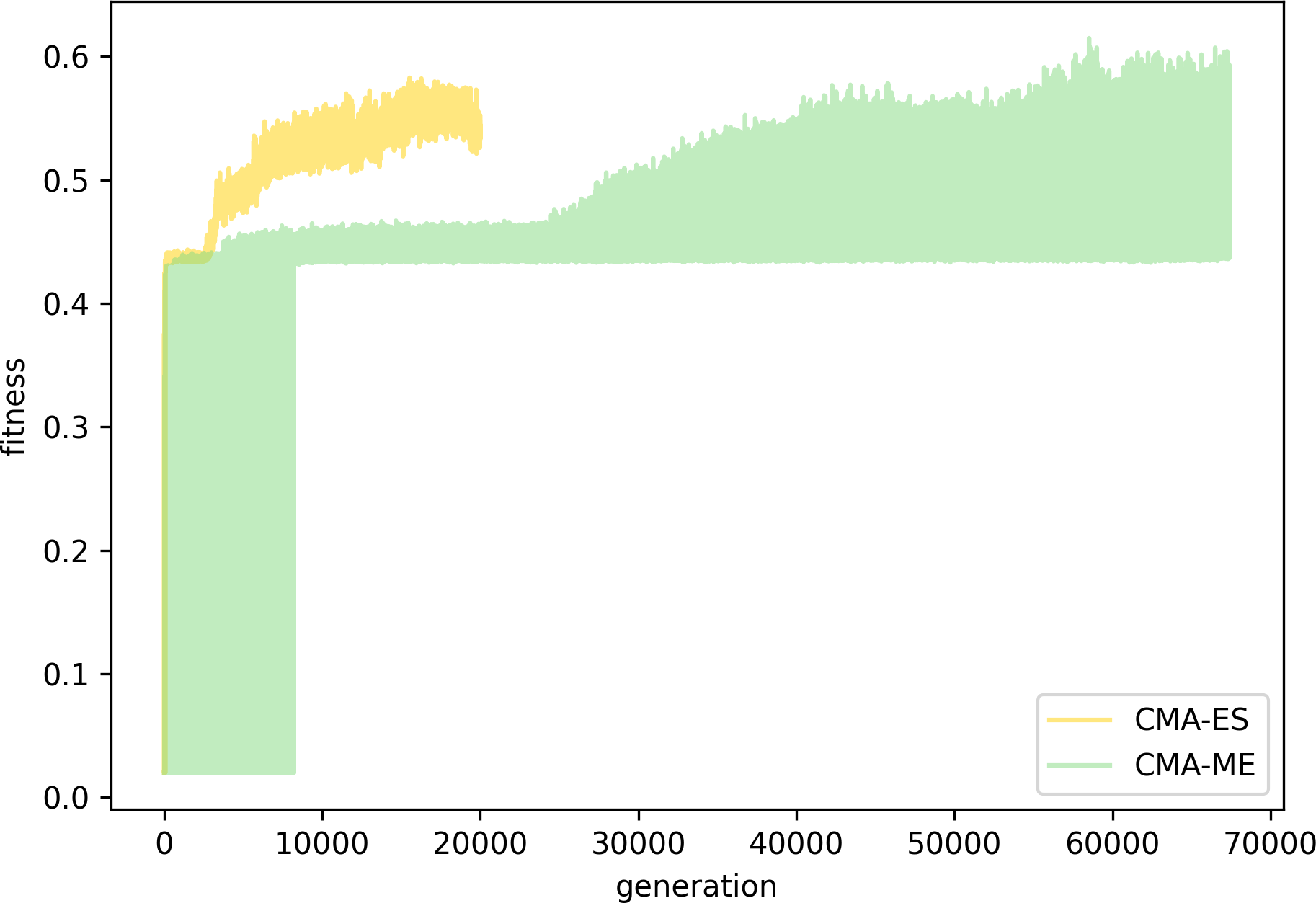}%
  }\hfill
  \subcaptionbox{Light chasing with obstacle\label{fig:loss_env2}}{%
    \includegraphics[width=0.3\textwidth]{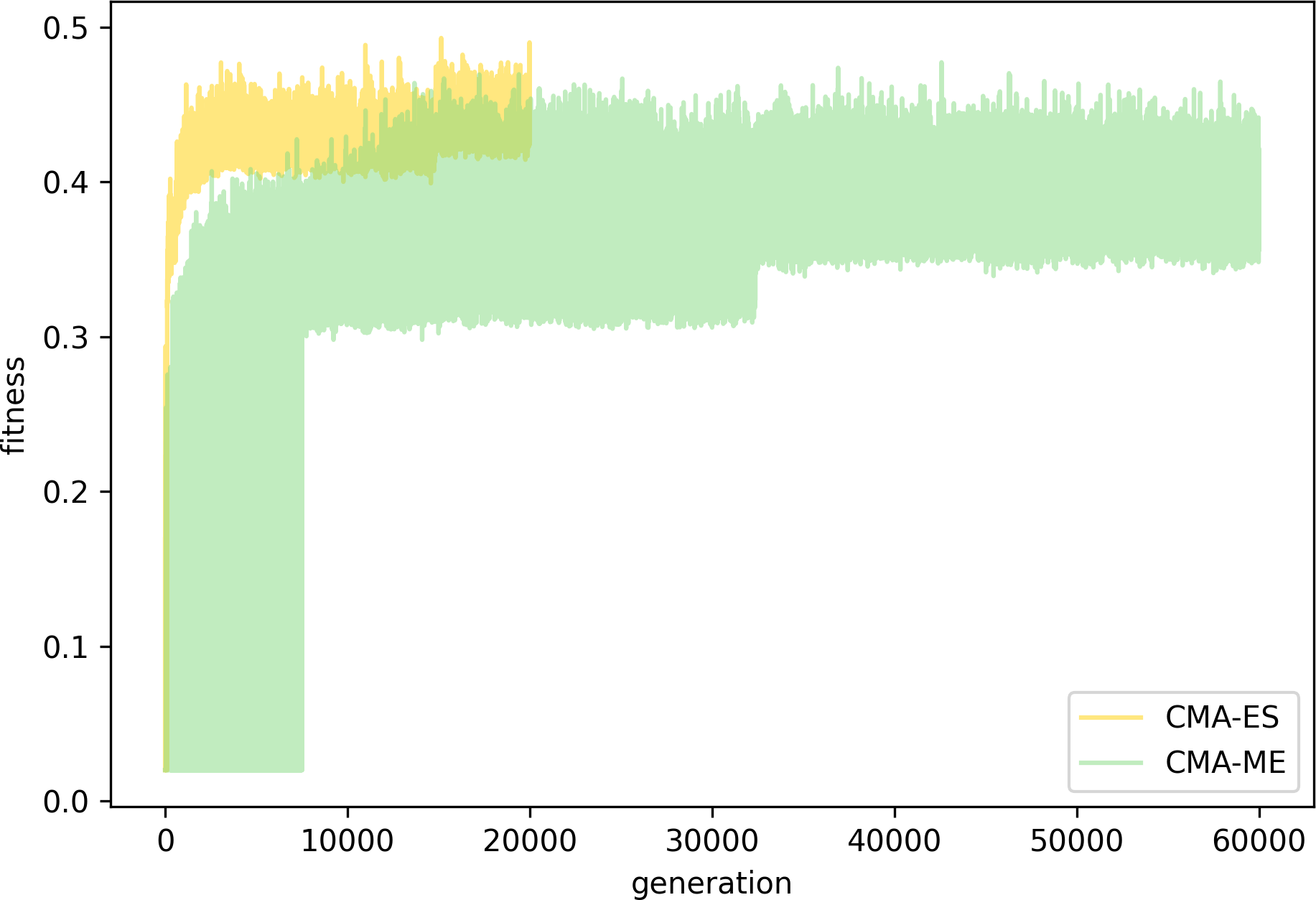}%
  }\hfill
  \subcaptionbox{Carrying ball to target\label{fig:loss_env3}}{%
    \includegraphics[width=0.3\textwidth]{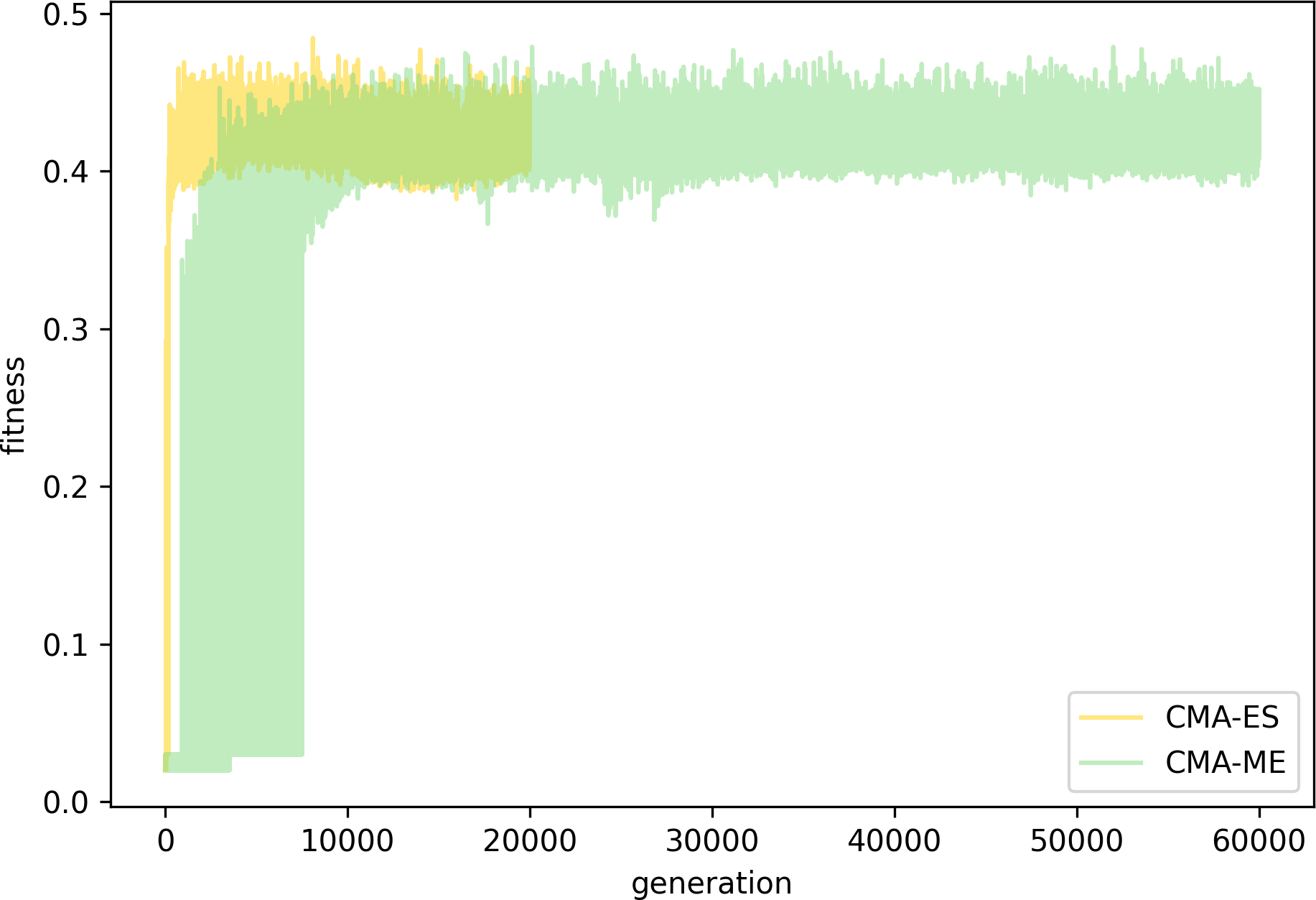}%
  }
  \caption{Maximum fitness score through generations.}
  \label{fig:loss}
\end{figure}

Fig.~\ref{fig:env_es_body} shows all robot designs with the best fitness scores in regards to their training method and task. Almost all robots for the LC and CBT tasks fill the entire $5\times 5$ grid of cells. Those environments do not have any environmental constraints (any obstacle) for the robot size. Therefore, we infer that the full grid of modules is easier to design and there are more computational resources for controlling the robot. Their fitness scores are shown in Table~\ref{tab:train_fitness}. The results indicate that CMA-ES and CMA-ME can reach almost the same fitness scores after training. However, CMA-ME has fewer generations for the 15 emitters (4,000 generations per emitter). It is possible that if we run 20,000 generations per emitter, CMA-ME could reach a better final performance than CMA-ES and with more diversity. The history of the maximum fitness score per generation is depicted in Fig.~\ref{fig:loss}.

\begin{table}
\centering
\caption{Best fitness score after training in the tasks of light chasing (LC), light chasing with obstacle (LCO) and carrying ball to target (CBT)}
\label{tab:train_fitness}
\begin{tabular}{c|c|c|}
\cline{2-3}
                           & CMA-ES  & CMA-ME  \\ \hline
\multicolumn{1}{|c|}{LC} & 0.58274 & 0.61481 \\ \hline
\multicolumn{1}{|c|}{LCO} & 0.49295 & 0.47723 \\ \hline
\multicolumn{1}{|c|}{CBT} & 0.48445 & 0.47884    \\ \hline
\end{tabular}
\end{table}

\begin{figure}[htb]
\centering
  \includegraphics[width=0.7\linewidth]{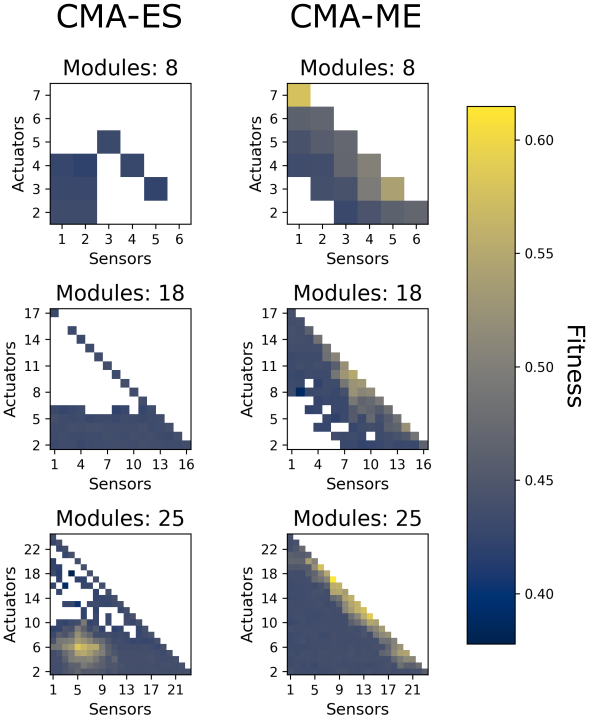}
  \caption{Selected elites trained in the light chasing environment. Those modules were selected because they are the most different between CMA-ES and CMA-ME. Axes and subplots indicate the number of components.}
  \label{fig:elites}
\end{figure}

The elites were saved for both CMA-ES and CMA-ME, then we can compare their quality diversity. In Table~\ref{tab:elites}, the number of cells filled and QD-scores of all six methods and tasks combinations are presented. It is noticeable that CMA-ME provides much more quality diversity because of its bigger number of feature configurations and its QD-score. We can visualize it in Fig.~\ref{fig:elites}. This shows a small part of the elites produced for the light chasing tasks with CMA-ES and CMA-ME. Nevertheless, it confirms those two quality diversity measurements because more cells are filled, and there are more cells with higher fitness scores.

\begin{table}
\centering
\caption{Elites stored during training for the light chasing (LC), light chasing with obstacle (LCO) and carrying ball to target (CBT)}
\label{tab:elites}
\begin{tabular}{c|cc|cc|}
\cline{2-5}
                          & \multicolumn{2}{c|}{CMA-ES}                  & \multicolumn{2}{c|}{CMA-ME}                  \\ \cline{2-5} 
                          & \multicolumn{1}{c|}{Cells filled} & QD-score & \multicolumn{1}{c|}{Cells filled} & QD-score \\ \hline
\multicolumn{1}{|c|}{LC}  & \multicolumn{1}{c|}{67.58\%}      & 0.29530  & \multicolumn{1}{c|}{89.57\%}      & 0.40152  \\ \hline
\multicolumn{1}{|c|}{LCO} & \multicolumn{1}{c|}{17.88\%}      & 0.06069  & \multicolumn{1}{c|}{61.80\%}      & 0.19841  \\ \hline
\multicolumn{1}{|c|}{CBT} & \multicolumn{1}{c|}{58.10\%}      & 0.22957  & \multicolumn{1}{c|}{93.82\%}      & 0.37996  \\ \hline
\end{tabular}
\end{table}

\begin{figure}[htpb]
\centering
  \subcaptionbox{CMA-ES - LC \#1\label{fig:env1_es_1}}{%
    \includegraphics[width=0.16\textwidth]{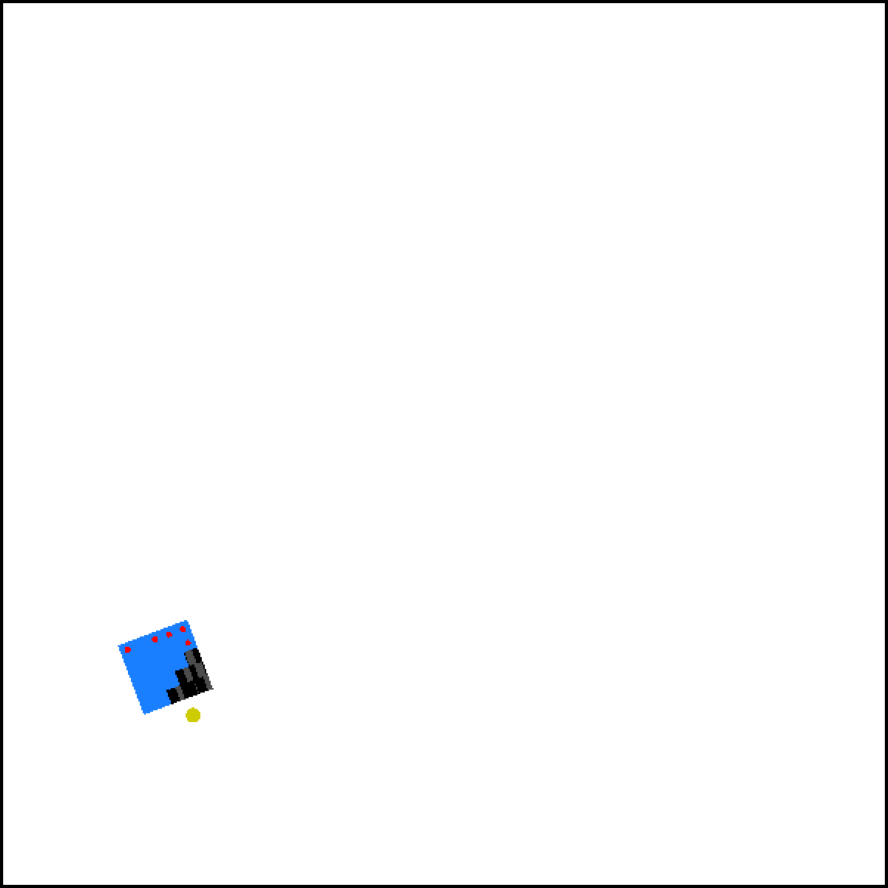}%
  }
  \subcaptionbox{CMA-ES - LC \#2\label{fig:env1_es_2}}{%
    \includegraphics[width=0.16\textwidth]{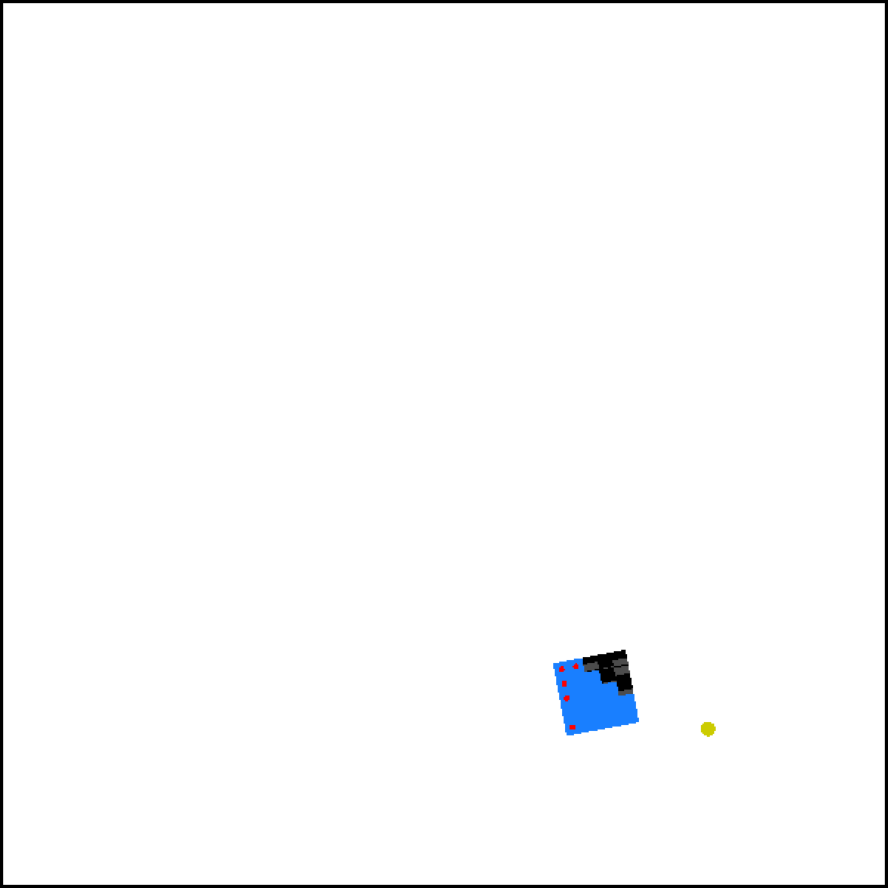}%
  }
  \subcaptionbox{CMA-ES - LC \#3\label{fig:env1_es_3}}{%
    \includegraphics[width=0.16\textwidth]{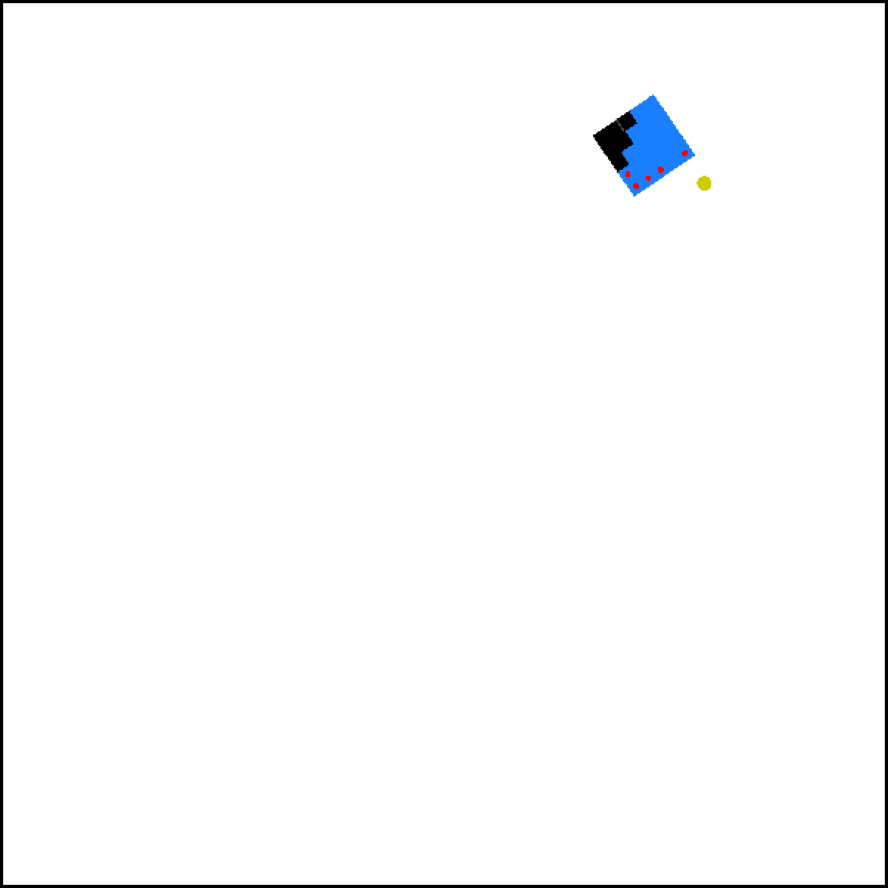}%
  }
  \subcaptionbox{CMA-ES - LC \#4\label{fig:env1_es_4}}{%
    \includegraphics[width=0.16\textwidth]{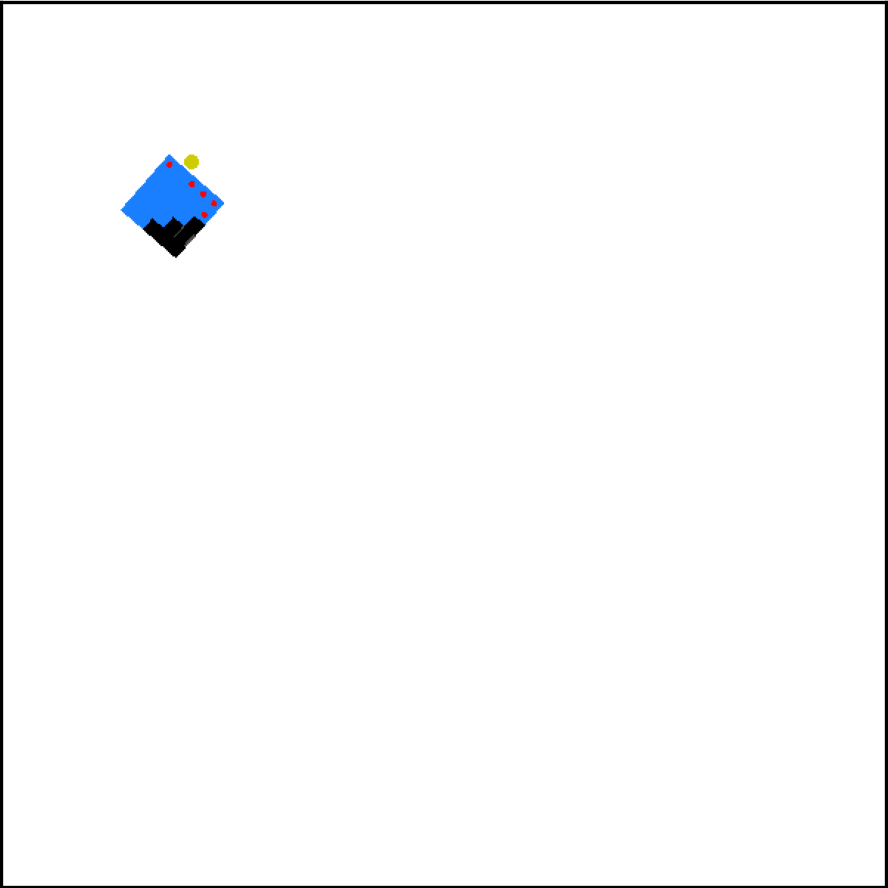}%
  }\\
  \subcaptionbox{CMA-ME - LC \#1\label{fig:env1_cmame_1}}{%
    \includegraphics[width=0.16\textwidth]{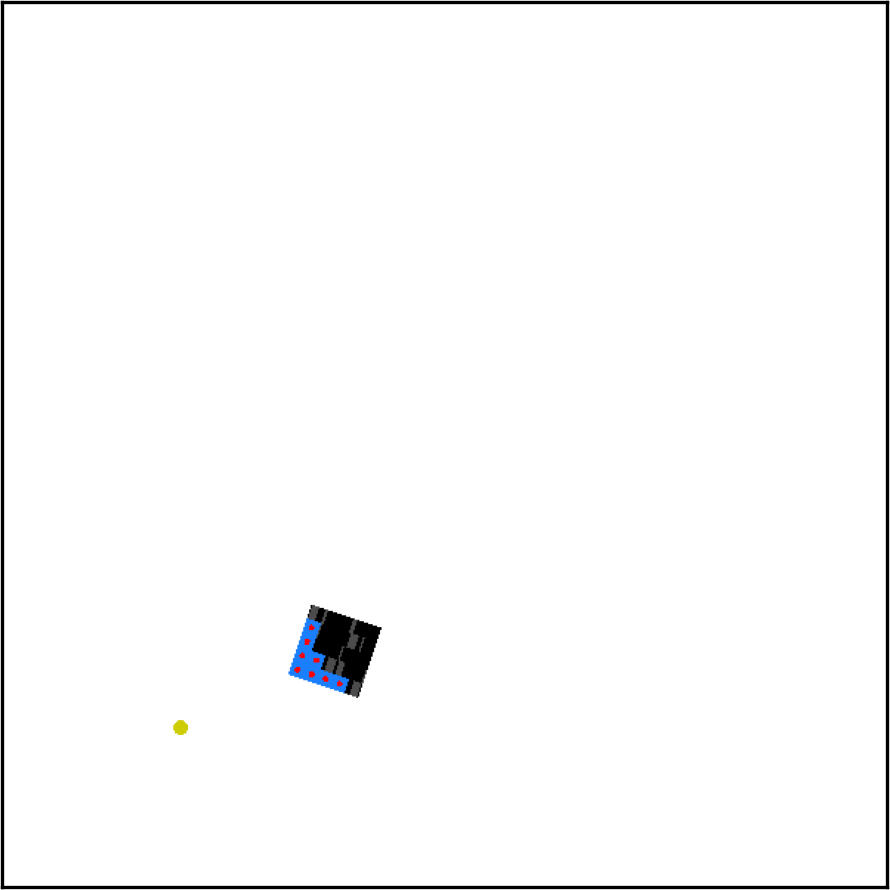}%
  }
  \subcaptionbox{CMA-ME - LC \#2\label{fig:env1_cmame_2}}{%
    \includegraphics[width=0.16\textwidth]{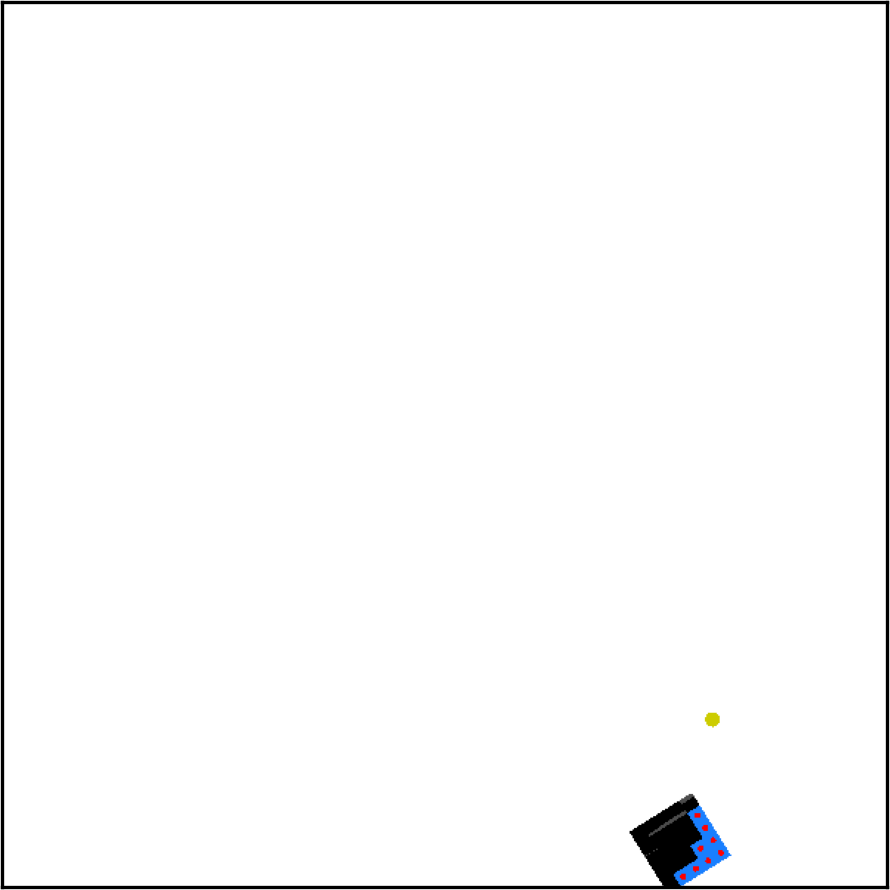}%
  }
  \subcaptionbox{CMA-ME - LC \#3\label{fig:env1_cmame_3}}{%
    \includegraphics[width=0.16\textwidth]{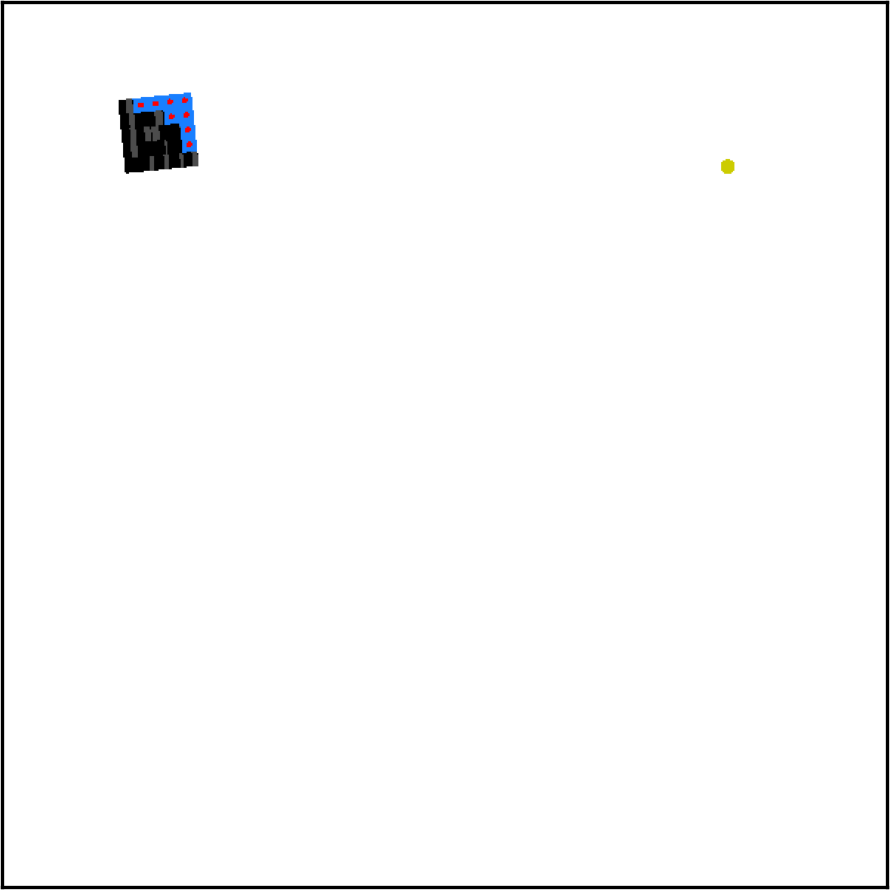}%
  }
  \subcaptionbox{CMA-ME - LC \#4\label{fig:env1_cmame_4}}{%
    \includegraphics[width=0.16\textwidth]{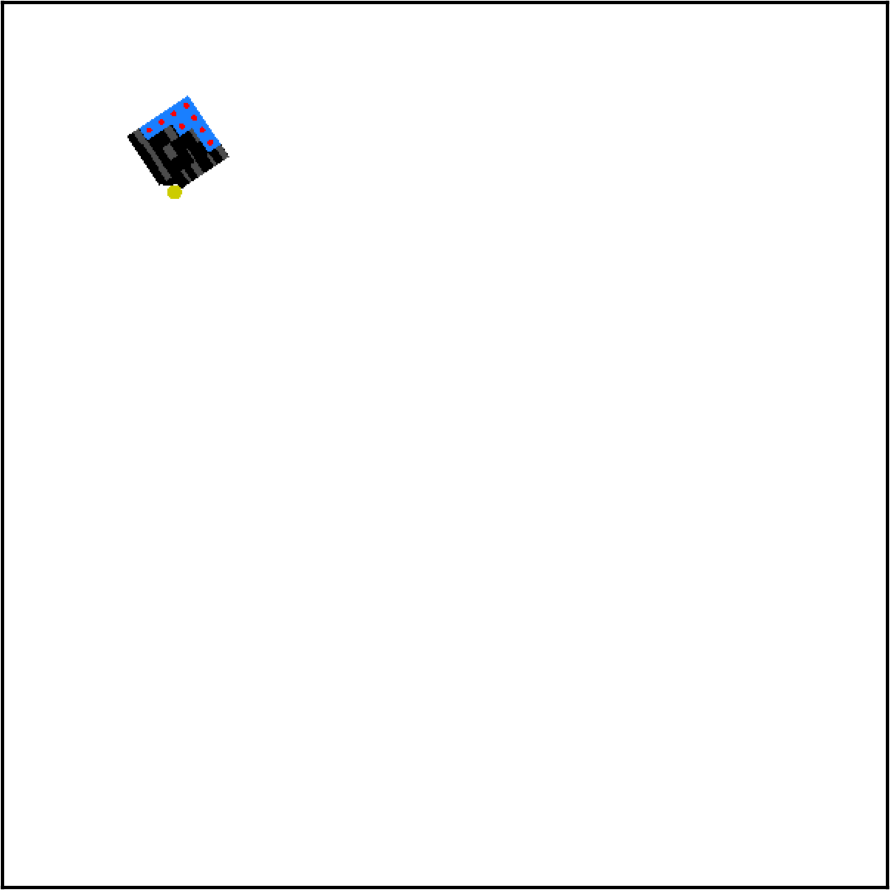}%
  }\\
    \subcaptionbox{CMA-ES - LCO \#1\label{fig:env2_es_1}}{%
    \includegraphics[width=0.16\textwidth]{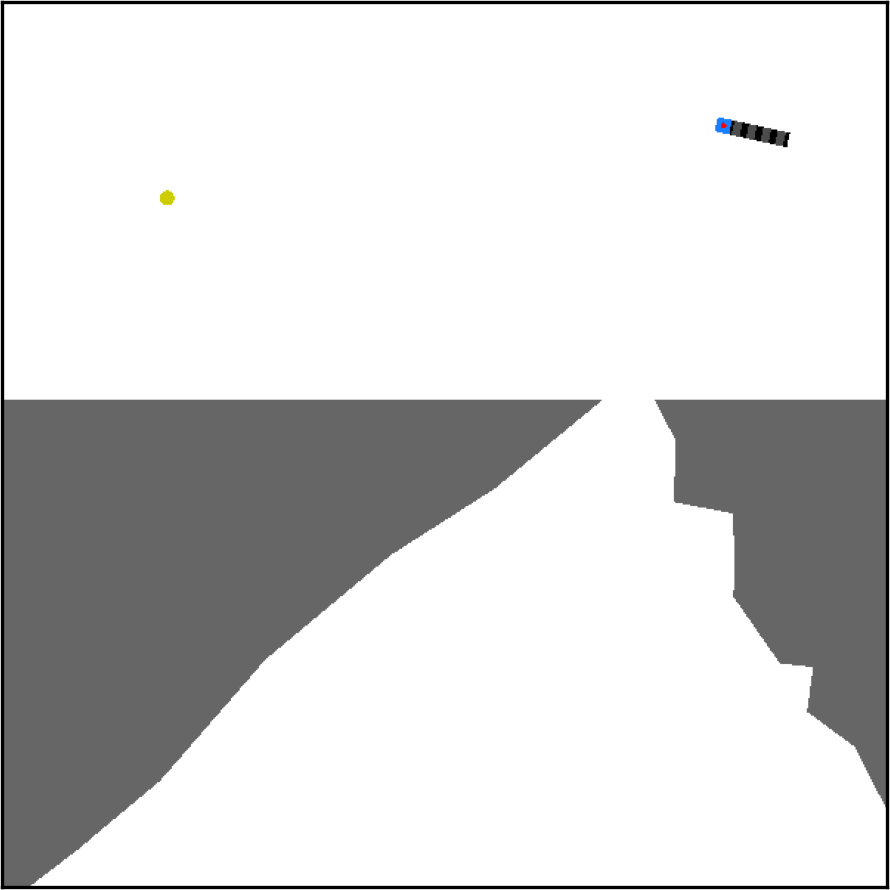}%
  }
  \subcaptionbox{CMA-ES - LCO \#2\label{fig:env2_es_2}}{%
    \includegraphics[width=0.16\textwidth]{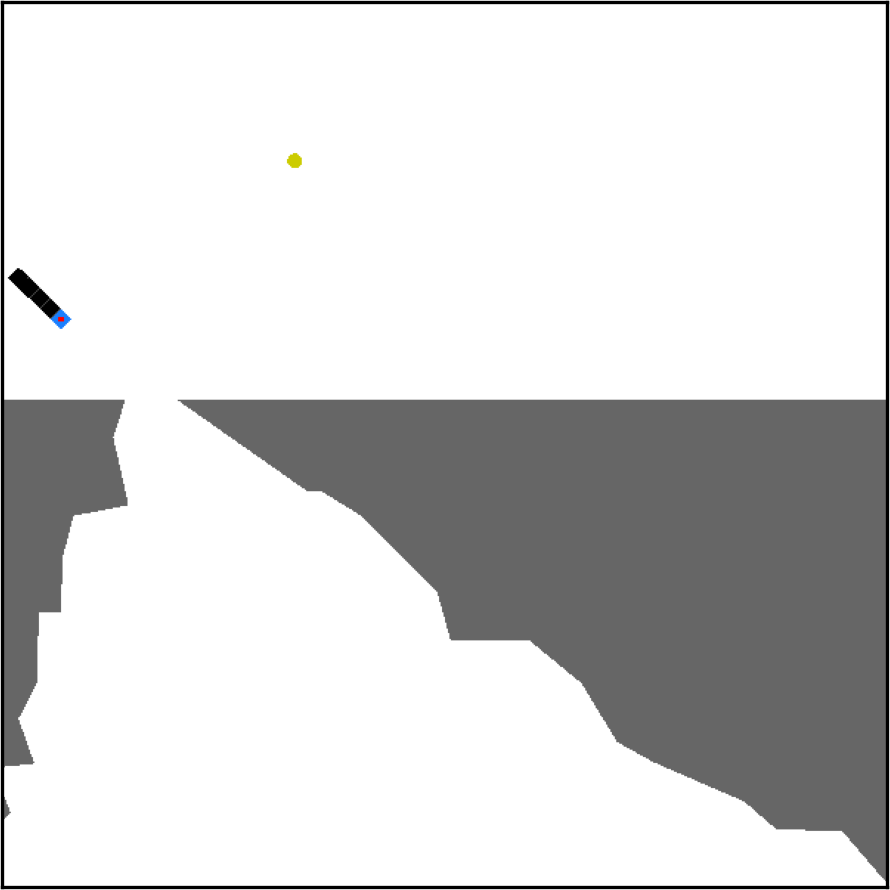}%
  }
  \subcaptionbox{CMA-ES - LCO \#3\label{fig:env2_es_3}}{%
    \includegraphics[width=0.16\textwidth]{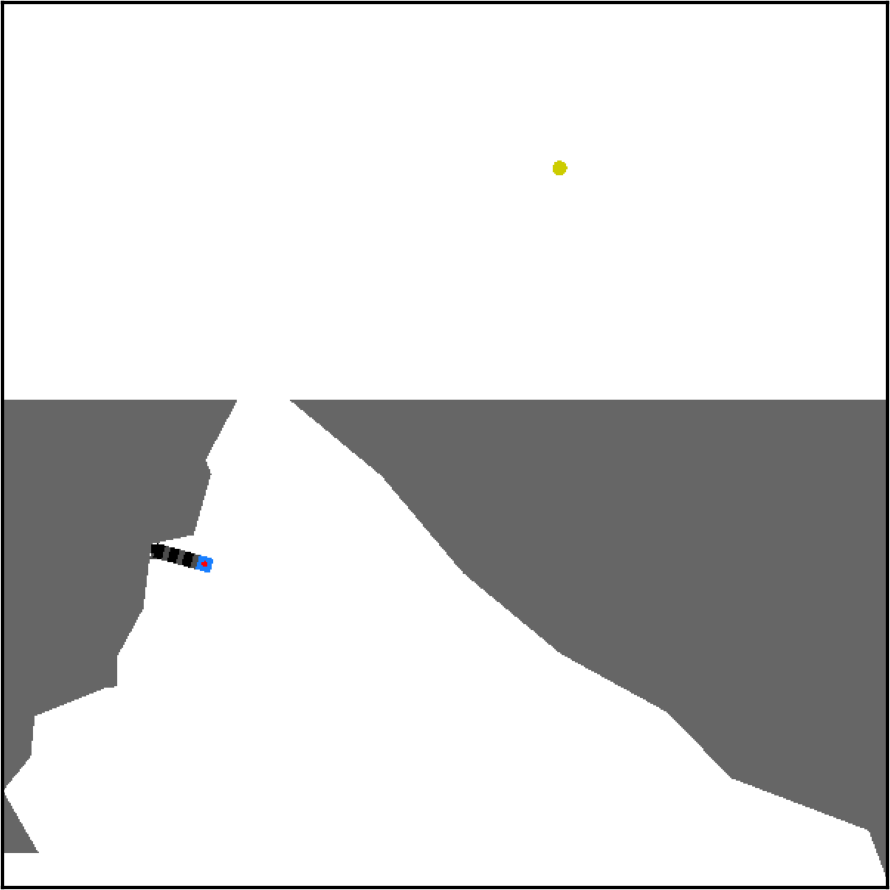}%
  }
  \subcaptionbox{CMA-ES - LCO \#4\label{fig:env2_es_4}}{%
    \includegraphics[width=0.16\textwidth]{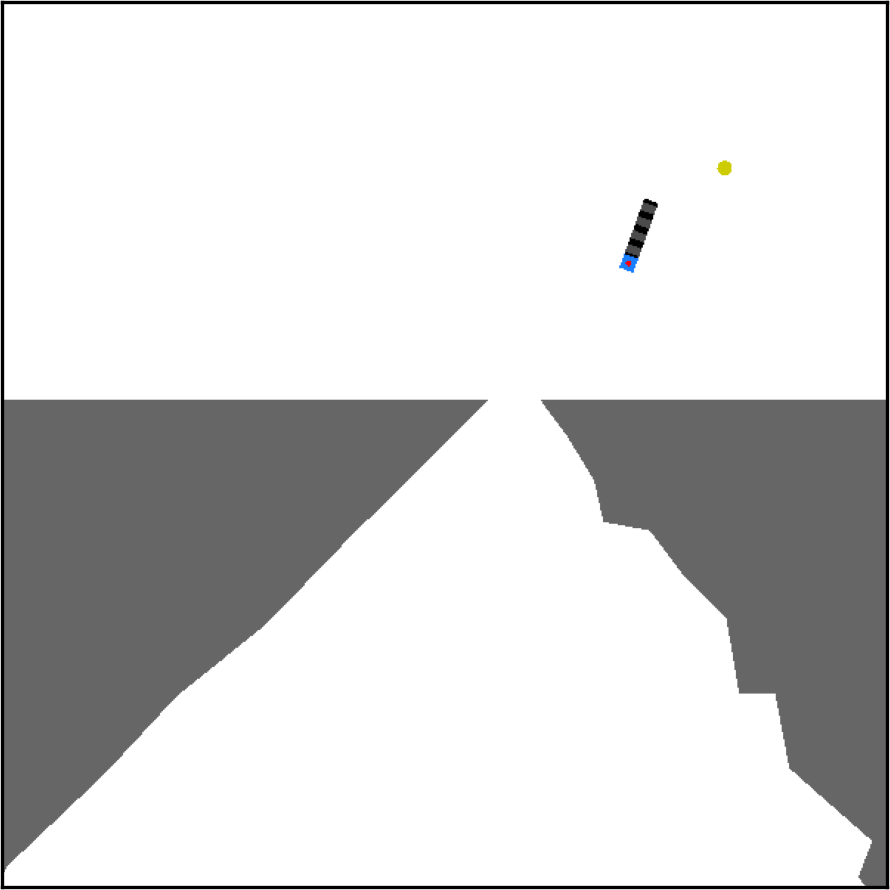}%
  }\\
     \subcaptionbox{CMA-ME - LCO \#1\label{fig:env2_cmame_1}}{%
    \includegraphics[width=0.16\textwidth]{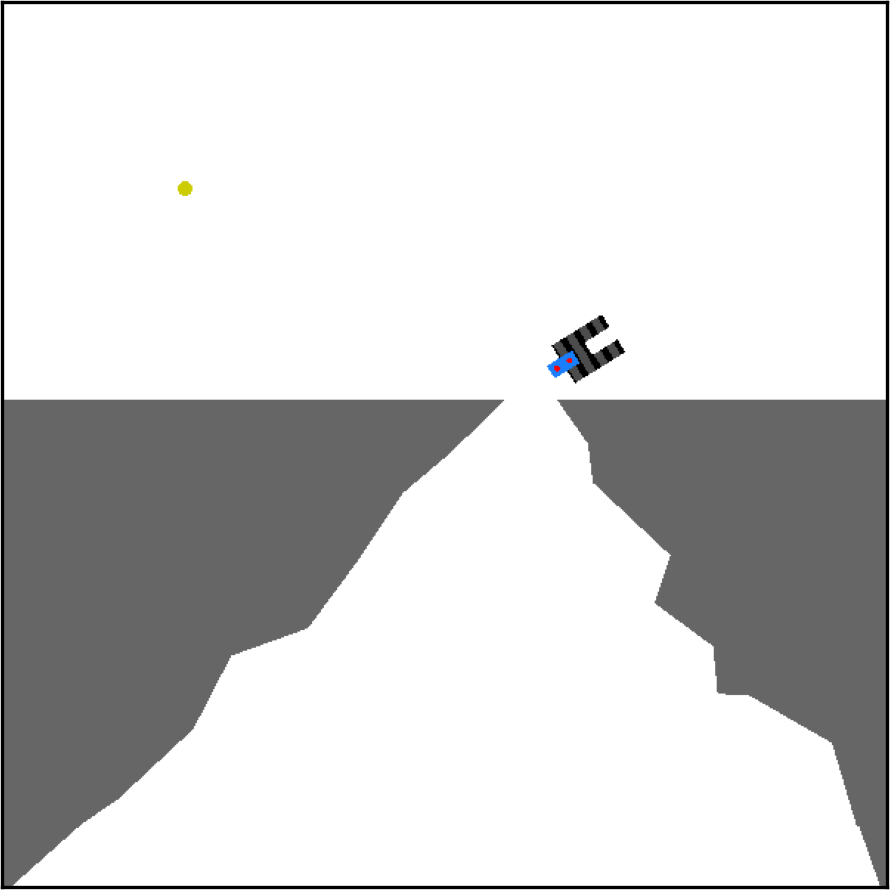}%
  }
  \subcaptionbox{CMA-ME - LCO \#2\label{fig:env2_cmame_2}}{%
    \includegraphics[width=0.16\textwidth]{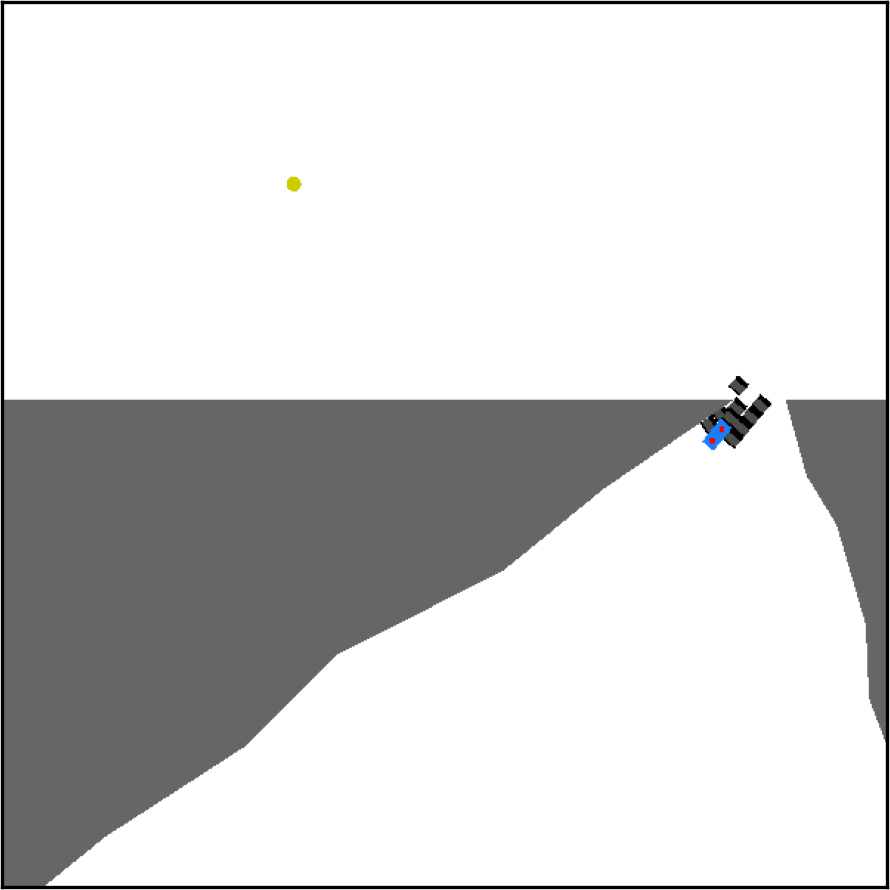}%
  }
  \subcaptionbox{CMA-ME - LCO \#3\label{fig:env2_cmame_3}}{%
    \includegraphics[width=0.16\textwidth]{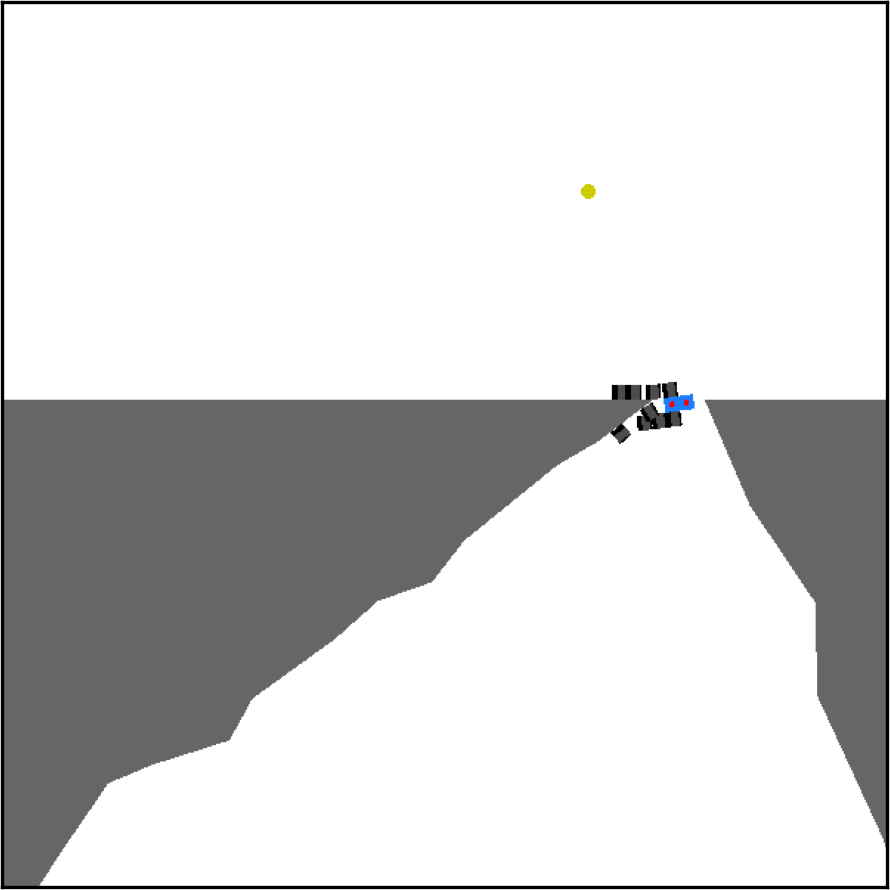}%
  }
  \subcaptionbox{CMA-ME - LCO \#4\label{fig:env2_cmame_4}}{%
    \includegraphics[width=0.16\textwidth]{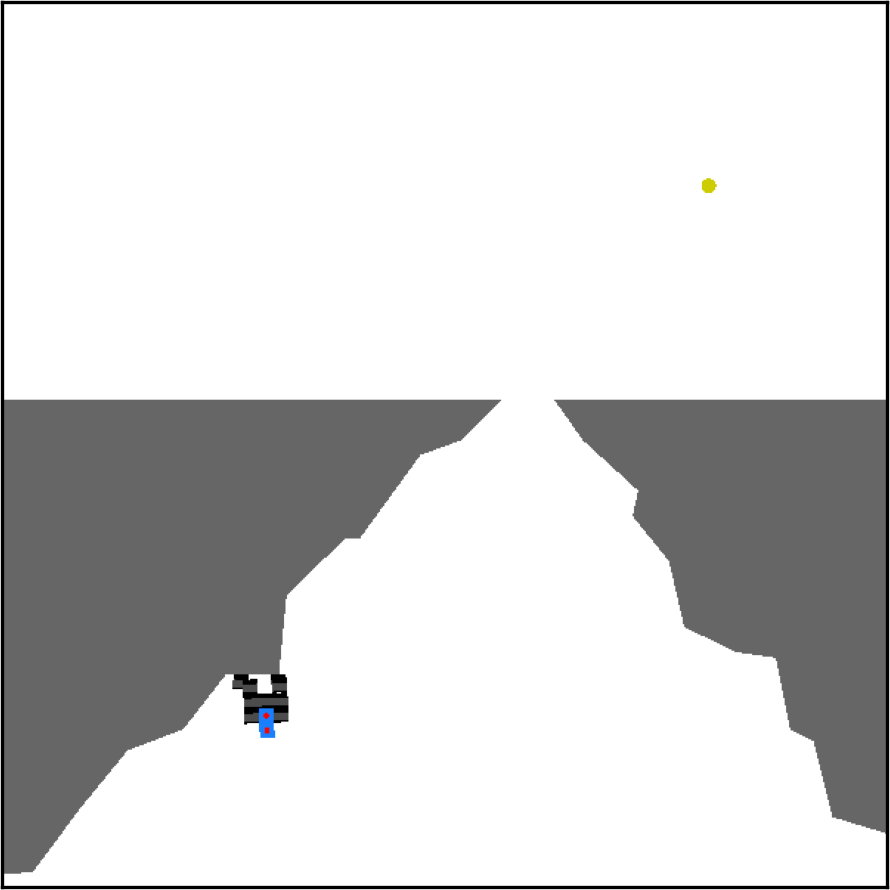}%
  }\\
    \subcaptionbox{CMA-ES - CBT \#1\label{fig:env3_es_1}}{%
    \includegraphics[width=0.16\textwidth]{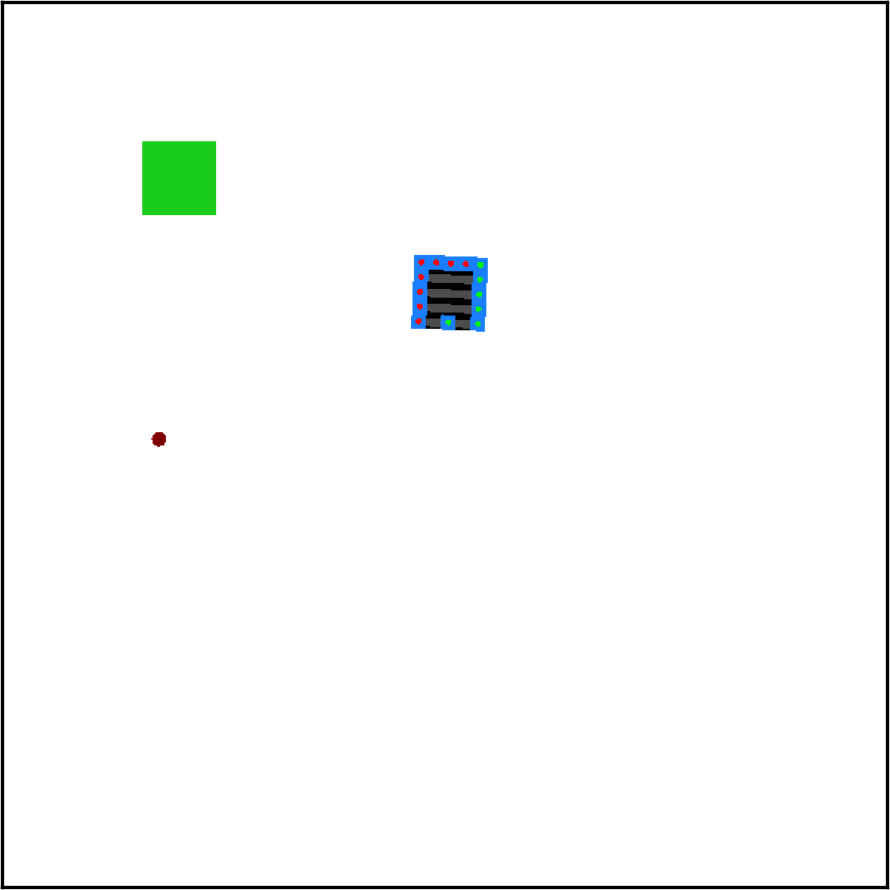}%
  }
  \subcaptionbox{CMA-ES - CBT \#2\label{fig:env3_es_2}}{%
    \includegraphics[width=0.16\textwidth]{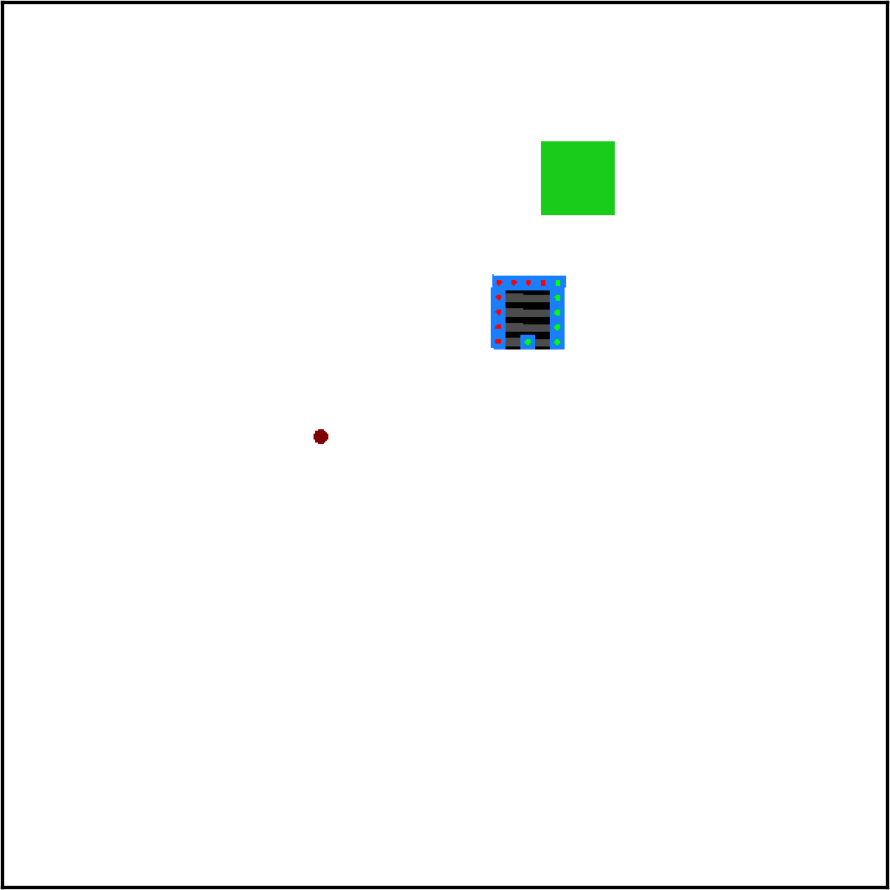}%
  }
  \subcaptionbox{CMA-ES - CBT \#3\label{fig:env3_es_3}}{%
    \includegraphics[width=0.16\textwidth]{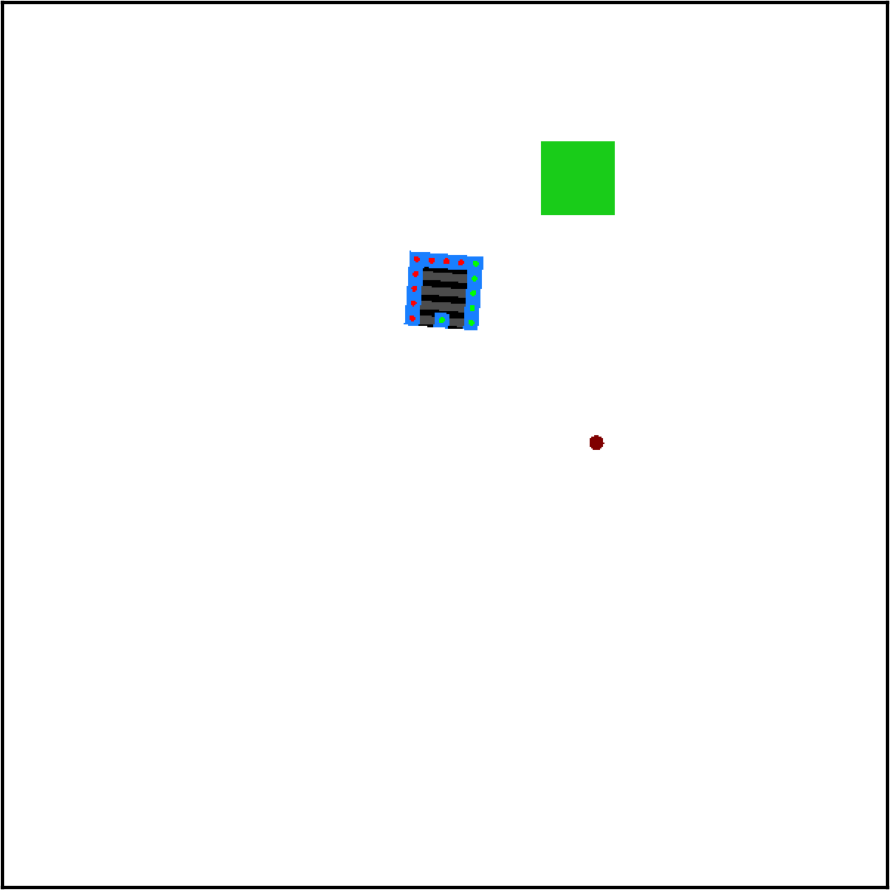}%
  }
  \subcaptionbox{CMA-ES - CBT \#4\label{fig:env3_es_4}}{%
    \includegraphics[width=0.16\textwidth]{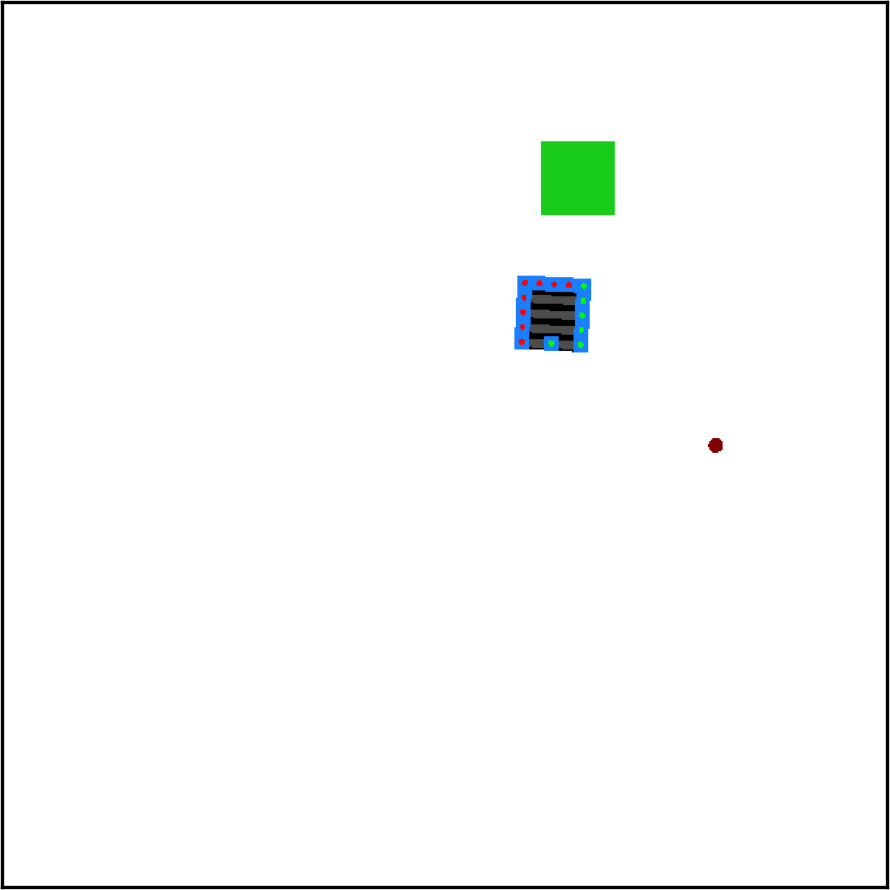}%
  }\\
    \subcaptionbox{CMA-ME - CBT \#1\label{fig:env3_cmame_1}}{%
    \includegraphics[width=0.16\textwidth]{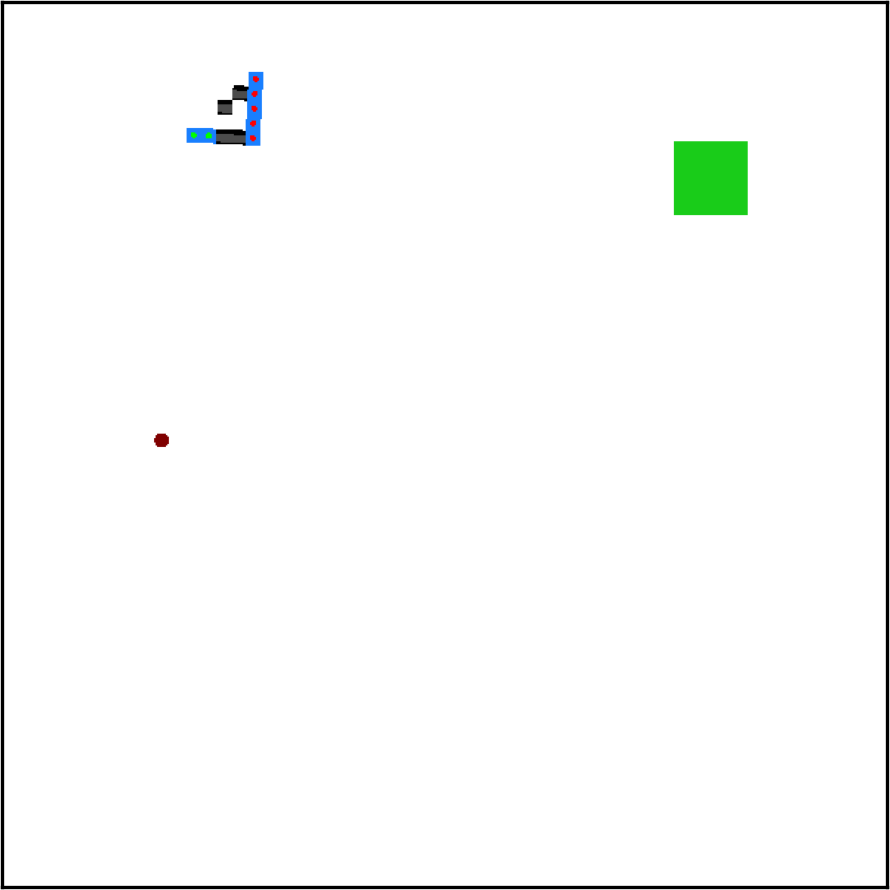}%
  }
  \subcaptionbox{CMA-ME - CBT \#2\label{fig:env3_cmame_2}}{%
    \includegraphics[width=0.16\textwidth]{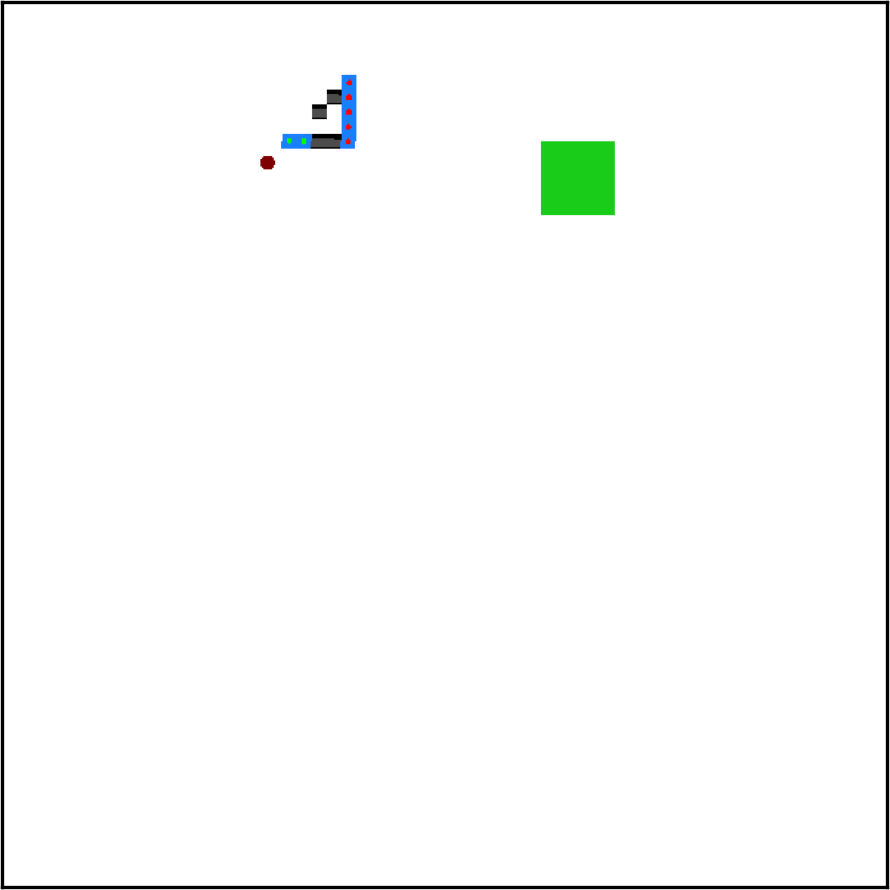}%
  }
  \subcaptionbox{CMA-ME - CBT \#3\label{fig:env3_cmame_3}}{%
    \includegraphics[width=0.16\textwidth]{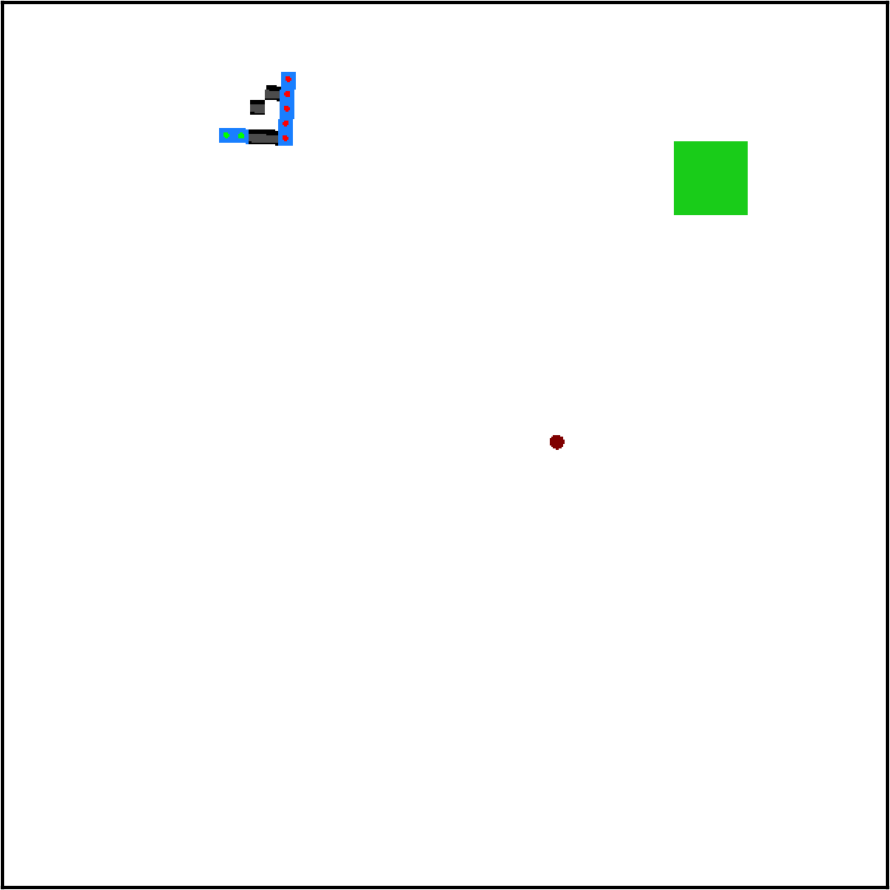}%
  }
  \subcaptionbox{CMA-ME - CBT \#4\label{fig:env3_cmame_4}}{%
    \includegraphics[width=0.16\textwidth]{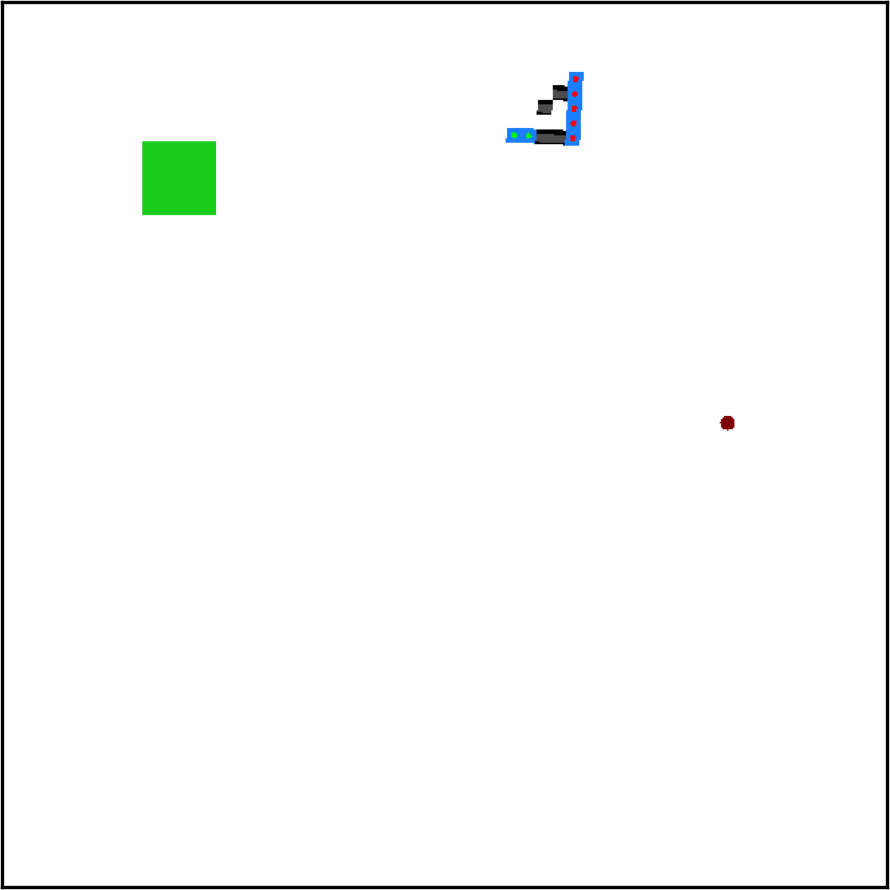}%
  }
  
  \caption{Last time-step where the robot is fully visible of the best NCRS trained with CMA-ES and CMA-ME in the environments for light chasing (LC), light chasing with obstacle (LCO) and carrying ball to target (CBT).}
  \label{fig:env_last_step}
\end{figure}

For testing the success of our six trained models, we run 100 times the simulation and the percentage of success is presented in Table~\ref{tab:success}. We can visualize some examples of those simulations in Fig.~\ref{fig:env_last_step}. The trained model with CMA-ES for the light chasing task got 92\% of success rate with $0.58274$ fitness score while the one trained with CMA-ME had 75\% of success and $0.61481$ of fitness score. This means that a higher fitness score does not indicate a more successful model for reaching the light. This can be observed in Fig.~\ref{fig:env1_es_1}-\subref{fig:env1_es_4} for CMA-ES, and Fig.~\ref{fig:env1_cmame_1}-\subref{fig:env1_cmame_4} for CMA-ME. We can see in Fig.~\ref{fig:env1_cmame_3} that the light is at the top-right corner and the robot goes to the top-left corner. This explains the 75\% success rate of this NCRS because the light is at the top-right corner in 25 out of the 100 simulations. This model learned to move faster to the light in the other three corners, but it misses the one in the top-right corner. For the light chasing with obstacle task, the reason for the higher success rate of CMA-ES robot is that it is much thinner than the CMA-ME robot. Therefore, it is easier to pass through the passage. If we define success in LCO by passing the center of the body through the passage, then CMA-ES and CMA-ME had a success rate, respectively, of 77\% and 45\%. The NCRS did not learn to move to the light after passing through the obstacle. It just moves forward. Because of the difficulty of this task, we can consider the results for LCO were partially successful in general and successful in body design. Fig.~\ref{fig:env2_es_1}-\subref{fig:env2_es_4} and Fig.~\ref{fig:env2_cmame_1}-\subref{fig:env2_cmame_4} show that. The task of carrying a ball to a target had no successful trained model. The robots for both training methods just move forward and, by chance, it moves the ball to target. This can be seen in Fig.~\ref{fig:env3_es_1}-\subref{fig:env3_es_4} and Fig.~\ref{fig:env3_cmame_1}-\subref{fig:env3_cmame_4}.

Fig.~\ref{fig:channel} shows how the channels progress through time. The hidden channels are predominantly different in their behavior for the developmental and control phases. We infer this is mainly due to the control flag channel which regulates these two phases. We can observe the different patterns that emerged in their final time-steps. From the initial "seed" state to the state in Fig.~\ref{fig:channel_2}, we can see how the NCA behaves during 10 time-steps of the developmental phase. In Fig~\ref{fig:channel_3}, we can see the end of the control phase during its 200 time-steps (100 time-steps in the environment). We can still understand its behavior because the hidden and input/output channels were set to zero at the beginning of the control phase, and the body, control flag, tissue, sensor, and actuator channels were fixed according to the morphology of the robot.

\begin{table}
\centering
\caption{Testing success percentage over 100 runs for the tasks of light chasing (LC), light chasing with obstacle (LCO) and carrying ball to target (CBT)}
\label{tab:success}
\begin{tabular}{c|c|c|}
\cline{2-3}
                           & CMA-ES  & CMA-ME  \\ \hline
\multicolumn{1}{|c|}{LC} & 92\% & 75\% \\ \hline
\multicolumn{1}{|c|}{LCO} & 20\% & 8\% \\ \hline
\multicolumn{1}{|c|}{CBT} & 1\% & 2\%        \\ \hline
\end{tabular}
\end{table}

\section{Discussion and conclusion}

Body-brain co-evolution is a challenging task \citep{bhatia2021evolution}. In this work, we developed three benchmark tasks for robot co-design and introduced a novel method by having a unified substrate as a genome with its own rules. This substrate is a single neural cellular automaton that works to develop and control a modular robot. This novelty opens up several possibilities in open-ended evolution \citep{stanley2019open}, especially because body and brain can co-evolve to the limits of the capacity of the artificial neural network. Because it defines the local rules in the CA, NCRS has the advantage of scalability. We also infer that curriculum learning will be important for complexifying the evolving robot \citep{bengio2009curriculum}. For example, the number of body parts and dimensions can increase over time with the progress of the generations. Evolution in multi-agent environments may also be applied, such as in PolyWorld \citep{yaeger1994computational}. We can also try to remove the two separated phases into one. Thus, we can observe how development and control can emerge and the performance the modular robots can have.

The presented results were successful for the LC task, but our trained models presented some failures when increasing the difficulty of the tasks. This may be addressed by adjusting the fitness score to reflect the success conditions, as well as by applying curriculum learning \citep{bengio2009curriculum}. In future works, we plan to apply our method in the Evolution Gym \citep{bhatia2021evolution}, or in a modified version of VoxCraft \citep{liu2020voxcraft} for 3D soft robots. Moreover, we aim at training and testing our approach for self-repair and robustness to noise.

\section*{Acknowledgment}

This work was partially funded by the Norwegian Research Council (NFR) through their IKTPLUSS research and innovation action under the project Socrates (grant agreement 270961). We thank Henrique Galvan Debarba for his thoughtful comments about the text. We also thank Joachim Winther Pedersen, Djordje Grbic, Miguel González Duque, and Rasmus Berg Palm for the helpful discussions during the implementation of the experiments.

\FloatBarrier
\bibliographystyle{plainnat}
\bibliography{references}

\end{document}